\pdfoutput=1

\documentclass[11pt]{article}

\usepackage[final]{acl}
\usepackage{times}
\usepackage{latexsym}
\usepackage{booktabs}
\usepackage{xcolor}
\usepackage{tcolorbox}
\usepackage{algorithm}
\usepackage{algpseudocode}
\usepackage{nicefrac}
\usepackage{color, colortbl}
\definecolor{Gray}{gray}{0.9}
\usepackage{multirow}
\usepackage{amsmath}
\usepackage{soul}
\usepackage{arydshln}  
\usepackage{enumitem}
\usepackage{siunitx} 
\usepackage{marginnote}
\usepackage{subcaption}
\usepackage{siunitx}
\usepackage{pifont}
\usepackage[T1]{fontenc}

\usepackage[utf8]{inputenc}

\usepackage{microtype}

\usepackage{inconsolata}

\usepackage{graphicx}

\title{AudioJudge: Understanding What Works\\ in Large Audio Model Based Speech Evaluation}

\author{
 \textbf{Potsawee Manakul\textsuperscript{1,3*}},
 \textbf{Woody Haosheng Gan\textsuperscript{2*}},
 \textbf{Michael J. Ryan\textsuperscript{3}},
 \textbf{Ali Sartaz Khan\textsuperscript{3}},
\\
 \textbf{Warit Sirichotedumrong\textsuperscript{1}},
 \textbf{Kunat Pipatanakul\textsuperscript{1}},
 \textbf{William Held\textsuperscript{3}},
 \textbf{Diyi Yang\textsuperscript{3}}
\\ 
 \textsuperscript{1}SCB 10X, SCBX Group,
 \textsuperscript{2}University of Southern California,
 \textsuperscript{3}Stanford University
}

\begin{document}
\maketitle
\def\thefootnote{*}\footnotetext{Equal contribution. Project code: \href{https://github.com/Woodygan/AudioJudge}{AudioJudge} \\Contacts: \href{mailto:potsawee@scb10x.com}{potsawee@scb10x.com}, \href{mailto:woodygan@usc.edu}{woodygan@usc.edu}}\def\thefootnote{\arabic{footnote}}

\begin{abstract}
Current speech evaluation suffers from two critical limitations: the need and difficulty of designing specialized systems targeting individual audio characteristics, and poor correlation between automatic evaluation methods and human preferences. This work presents a systematic study of Large Audio Model (LAM) as a Judge, AudioJudge, investigating whether it can provide a unified evaluation framework that addresses both challenges. We systematically explore AudioJudge across audio characteristic detection tasks, including pronunciation, speaking rate, speaker identification and speech quality, and system-level human preference simulation for automated benchmarking. We investigate different prompt engineering strategies, finding that audio concatenation combined with in-context learning significantly improves performance across both audio characteristic detection and human preference simulation tasks. We further introduce a multi-aspect ensemble AudioJudge to enable general-purpose multi-aspect audio evaluation. This method decomposes speech assessment into specialized judges for lexical content, speech quality, and paralinguistic features, achieving up to 0.91 Spearman correlation with human preferences on our system ranking benchmark. Robustness analysis reveals that while LAMs maintain strong performance under acoustic noise, they exhibit significant verbosity and positional biases that require careful mitigation.
\end{abstract}

\section{Introduction}

Current speech evaluation paradigms suffer from two critical limitations that hinder the development and comparison of speech generation systems: (1) {Speech evaluation typically requires specialized systems targeting individual audio characteristics.} Practitioners have used separately trained models for tasks like speech quality~\citep{utmos_interspeech, mosanetplus_icme} and pronunciation evaluation~\citep{deseyssel2023emphassess}. Each system can demand costly training or customization, creating barriers to broad evaluation tasks. (2) While static benchmarks often fail to capture the nuanced quality judgments that users make in speech-based applications~\citep{li2025mind}, manual evaluations can be too costly. Current automatic benchmarking for speech-in speech-out systems achieves low consistency with human judgments \citep{jiang2025s2s}.

Large Audio Models (LAMs) that process speech and generate natural language responses~\citep{tang2024salmonn, qwen2audio, held2024distilling} presents an opportunity to address both challenges through a unified evaluation framework. Prompting LAMs to act as judges (referred to as AudioJudge in this work), analogous to the successful LLM-as-a-Judge paradigm for text evaluation~\citep{zheng2023judging}, can potentially reduce the need for specialized model training and simulates human preferences well.

While prior works have used AudioJudge~\citep{yang2025emotts, chen2025audio}, they have focused on individual tasks and have not explored how to best prompt LAMs for AudioJudge pipelines. In order to provide guidance of when and how to effectively use AudioJudge, we present a systematic study that explores the use of AudioJudge across two evaluation scenarios that target distinct use cases mentioned above:
(1) \textbf{Audio characteristic detection} that targets practitioners who need to assess specific speech properties—pronunciation accuracy, speaking rate, speaker identification, and speech quality \textit{at the example level}. This capability enables rapid analysis of generated speech and audio recordings without requiring specialized trained models for each characteristic.
(2) \textbf{Overall human preference correlation} for evaluating how well LAMs can replicate human preferences when ranking speech generation systems \textit{at the system level}. This is useful for automated tools to evaluate and compare speech generation systems.

We investigate the design space of AudioJudge for speech evaluation, exploring how different prompting strategies affect performance across diverse audio characteristics. Additionally, we conduct comprehensive robustness analyses on AudioJudge, examining its behavior when subject to noise, its susceptibility to verbosity and positional biases, which are limitations previously observed in LLM-as-a-Judge~\citep{saito2023verbosity, zheng2023judging}. We present both strengths and limitations of current LAM evaluation capabilities.
In summary, this work makes the following contributions:
\begin{enumerate}[leftmargin=*]
\setlength\itemsep{-0.2em}
    \item We provide a systematic study of \textit{AudioJudge} across diverse speech evaluation tasks; demonstrating its strengths and weaknesses for evaluating both example-level audio characteristics and system-level performance.
    
    \item We investigate prompt engineering strategies tailored for audio evaluation, introducing audio concatenation techniques that improve performance at both example and system levels, and a multi-aspect ensemble approach that improves correlation with human judgment.
    
    \item We conduct thorough robustness checks finding that LAMs maintain stability under acoustic noise, but exhibit significant verbosity and positional biases that require mitigation.
\end{enumerate}

\section{Related Work}
\paragraph{LLM-as-a-Judge for Multimodal Evaluation.} 
The success of LLM-as-a-Judge for text evaluation \citep{zheng2023judging, dubois2023alpacafarm} has inspired extensions to vision~\citep{xiong2024llavacritic, chen2024unified} and audio. While some prior works apply text-based LLMs to transcripts of speech~\citep{latif2023emotionchatgpt, efstathiadis2025llm, yang2023generative}, the most closely related research to ours centers on using AudioJudge models primarily for assessing speech quality~\citep{deshmukh2024pam, wang2024enabling, chen2025audio}. In contrast, we aim to investigate whether a single model can reliably evaluate a broad range of dimensions which practitioners might evaluate.

\paragraph{Specialized Speech Evaluation Systems.} Traditional speech evaluation often relies on specialized models: UTMOS, MOSANET+, MOSNet, and DNSMOS for speech quality~\citep{utmos_interspeech,mosanetplus_icme,lo2019mosnet, reddy2021dnsmos}, STOI and NISQA for intelligibility~\citep{taal2011algorithm,mittag2021nisqa}, and task-specific systems for prosody~\citep{deseyssel2023emphassess} and pronunciation~\citep{deseyssel2023emphassess, korzekwa2021weakly}. Some toolkits, such as VERSA~\citep{shi-etal-2025-versa} combine many aspect-specific metricsfor comprehensive speech quality analysis. While effective within domains, developing each metric or model requires extensive labeled data and custom architectures, creating scalability barriers that AudioJudge can address.

\paragraph{Speech Benchmarking and Human Preferences.} Existing benchmarks evaluate speech systems across tasks: VoiceBench~\citep{chen2024voicebench} assesses general voice capabilities, SD-Eval~\citep{ao2024sdeval} measures speech understanding, MMAU~\citep{sakshi2024mmaumassivemultitaskaudio} evaluates multimodal audio understanding, AIR-Bench~\citep{yang2024air} tests comprehensive audio reasoning, AudioBench~\citep{wang2024audiobench} benchmarks audio-language models, SUPERB~\citep{yang2021superb} evaluates traditional speech processing, and SLURP~\citep{bastianelli2020slurp} measures spoken language understanding. However, these benchmarks measure objective metrics rather than capturing subjective human preferences. Prior work on human evaluation of speech systemsshows that static benchmarks poorly predict human preferences~\citep{li2025mind} and has concluded that LAMs are not straightforwardly usable for automatic evaluation~\citep{jiang2025s2s}. This work will evaluate AudioJudge on a range of tasks, and provide simple modifications to improve the correlation between automatic rankings and human preferences.

\section{Designing AudioJudge}
\label{section:designing_lam}

\subsection{AudioJudge Framework}
\label{subsec:lam_framework}

\begin{figure*}[!t]
    \centering
    \includegraphics[width=0.95\linewidth]{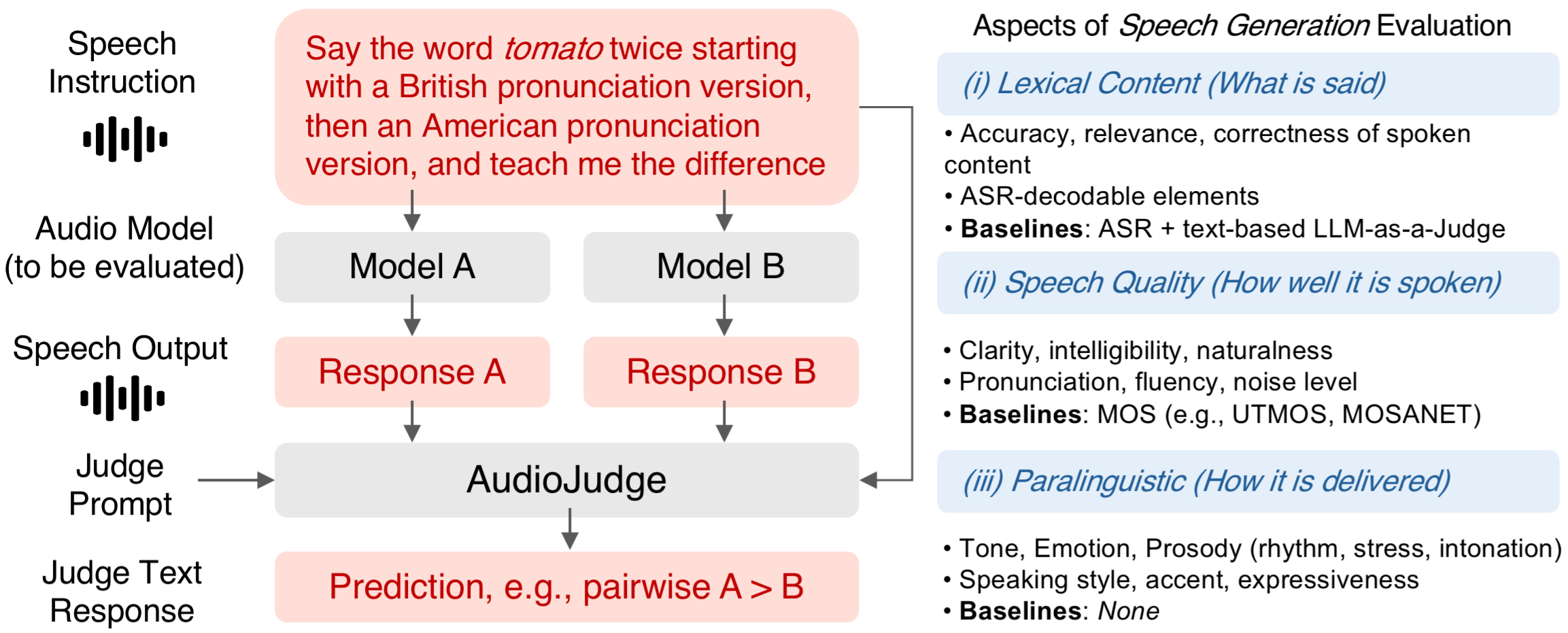}
    \caption{AudioJudge takes an instruction and audio responses and then performs evaluation (e.g., pairwise in this illustration) based on a judge evaluation prompt.}
    \label{fig:audiojudge}
\end{figure*}

AudioJudge prompts a large audio model (LAM) to act as a judge for speech evaluation tasks. Similar to LLM-as-a-Judge \citep{zheng2023judging, liusie-etal-2024-llm}, this framework can be implemented in multiple modes: (1) pointwise scoring, (2) reference-based comparison, and (3) pairwise comparison. 
In this work, we focus specifically on pairwise comparison, where the LAM directly compares two audio responses to determine which is better or whether they match in a certain way\footnote{In Appendix~\ref{appendix:pointwise_experiment},  we find that pairwise evaluation provides consistently more reliable results than pointwise evaluation.}. The prompting process for our pairwise comparison is visualized in Figure~\ref{fig:audiojudge}.

Prior work using on speech quality evaluation \citep{wang2024enabling, chen2025audio} has not explored how prompt engineering can improve performance. As such, we investigate several design strategies to enhance LAM evaluation capabilities across different types of speech evaluation.

\subsection{In-Context Audio Concatenation}
\label{ssec:icl}

\textit{In-Context Learning} (ICL) has proven effective for text-based LLMs~\citep{brown2020language, dong2022survey}, making it a natural candidate for exploration with \textit{AudioJudge} models. However, unlike text, audio inputs introduce unique challenges: multiple audio segments can be appended to a language-audio model’s (LAM’s) context in two primary ways---\textbf{naively}, where each audio segment is uploaded as a separate file, or via \textbf{concatenation}, where segments are merged into a single continuous audio stream with acoustic cues (e.g., pauses or boundary tones) between segments.

We investigate two key design dimensions (i) whether to concatenate in-context examples into longer audio sequences, and (2) whether to concatenate audio from test audio---response pairs. This yields five concatenation strategies: \textit{No Concatenation}, \textit{Pair Example Concatenation},  \textit{Examples Concatenation}, \textit{Test Concatenation}, and \textit{Examples \& Test Concatenation}.

The underlying intuition is that LAMs may better comprehend continuous audio streams than fragmented, alternating audio--text contexts. Detailed examples and prompt templates for each strategy are provided in Appendix~\ref{appendix:concatenation_prompt}.

\subsection{Transcript Information Augmentation}
We hypothesize that providing LAMs with additional textual information when processing audio might boost their performance. Specifically, we examine whether providing ground truth or ASR transcripts (GPT-4o Transcribe~\citep{openai2024gpt4otranscribe}) together with audio helps the model by releasing it from speech recognition and reducing the reasoning steps required for the evaluation task.

\subsection{Multi-Aspect Judge Ensemble}
\label{ssec:multi_aspect_ensemble}

For comprehensive speech evaluation tasks that involve judgment across multiple dimensions, we investigate whether decomposing the evaluation into individual aspects and then ensembling improves performance. Specifically, we introduce a multi-aspect ensemble approach with three judges differing by prompts and majority voting. We investigate this ensemble method on SpeakBench, which is a multi-aspect evaluation dataset in Section~\ref{section:system_level}. The three specialized judges are as follows:
\begin{itemize}[leftmargin=*]
\setlength\itemsep{-0.2em}
    \item \textbf{Lexical Judge}: Evaluates textual content (i.e., accuracy, completeness, organization) while ignoring audio qualities.
    \item \textbf{Paralinguistic Judge}: Assesses whether tone, prosody, expressiveness, and accent patterns satisfy the instruction's requirements while ignoring content quality.
    \item \textbf{Speech Quality Judge}: Focuses on clarity, naturalness, fluency, and pronunciation correctness while ignoring content and expressive features.
\end{itemize}

\noindent Each judge (differing in their prompts) independently produces a prediction (Audio 1 better, Audio 2 better, or tie) for the same audio pair. We then apply majority voting to determine the final ensemble prediction. Different prompting strategies (described in Section~\ref{ssec:icl}) can be used in the ensemble method. Section~\ref{section:fine_grained} compares these prompting strategies, and Section~\ref{section:system_level} applies the best-performing setup to the ensemble method.

\begin{table*}[!t]
  \small
  \tabcolsep=1.8mm
  \centering
  \begin{tabular}{l c c c c c c c c}
    \toprule
    \multirow{3}{*}{\textbf{Method}} & \multirow{3}{*}{\textbf{N-shot}} & \multicolumn{6}{c}{\textbf{Audio Characteristic Evaluation (Accuracy \%)}} & \multicolumn{1}{c}{\textbf{Average}} \\
    \cmidrule(lr){3-8}
     & & \multicolumn{3}{c}{\textbf{Paralinguistic}} & \multicolumn{3}{c}{\textbf{Speech Quality}} & \\
    \cmidrule(lr){3-5} \cmidrule(lr){6-8}
     & & \textbf{Prn} & \textbf{Speed} & \textbf{SkID} & \textbf{SOM} & \textbf{TMH} & \textbf{ThaM} & \\
    \midrule
    Random Guess         & N/A & \phantom{\textsuperscript{***}}50.0\phantom{\textsuperscript{***}} & \phantom{\textsuperscript{***}}50.0\phantom{\textsuperscript{***}} & \phantom{\textsuperscript{***}}50.0\phantom{\textsuperscript{***}} & \phantom{\textsuperscript{***}}50.0\phantom{\textsuperscript{***}} & \phantom{\textsuperscript{***}}50.0\phantom{\textsuperscript{***}} & \phantom{\textsuperscript{***}}50.0\phantom{\textsuperscript{***}} & \phantom{\textsuperscript{***}}50.0\phantom{\textsuperscript{***}} \\
    Baseline             & 0 & \phantom{\textsuperscript{***}}46.0\phantom{\textsuperscript{***}} & \phantom{\textsuperscript{***}}46.9\phantom{\textsuperscript{***}} & \phantom{\textsuperscript{***}}61.5\phantom{\textsuperscript{***}} & \phantom{\textsuperscript{***}}70.5\phantom{\textsuperscript{***}} & \phantom{\textsuperscript{***}}70.5\phantom{\textsuperscript{***}} & \phantom{\textsuperscript{***}}\textbf{65.5}\phantom{\textsuperscript{***}} & \phantom{\textsuperscript{***}}60.2\phantom{\textsuperscript{***}} \\
    + ASR Transcript     & 0 & \phantom{\textsuperscript{***}}46.5\phantom{\textsuperscript{***}} & \phantom{\textsuperscript{***}}41.9\phantom{\textsuperscript{***}} & \phantom{\textsuperscript{***}}53.5\phantom{\textsuperscript{***}} & \phantom{\textsuperscript{***}}72.5\phantom{\textsuperscript{***}} & \phantom{\textsuperscript{***}}\underline{71.5}\phantom{\textsuperscript{***}} & \phantom{\textsuperscript{***}}\underline{65.0}\phantom{\textsuperscript{***}} & \phantom{\textsuperscript{***}}58.5\phantom{\textsuperscript{***}} \\
    + Ground Truth       & 0 & \phantom{\textsuperscript{***}}63.0\textsuperscript{**}\phantom{\textsuperscript{*}} & \phantom{\textsuperscript{***}}39.1\phantom{\textsuperscript{***}} & \phantom{\textsuperscript{***}}51.8\phantom{\textsuperscript{***}} & \phantom{\textsuperscript{***}}70.0\phantom{\textsuperscript{***}} & \phantom{\textsuperscript{***}}66.0\phantom{\textsuperscript{***}} & \phantom{\textsuperscript{***}}64.5\phantom{\textsuperscript{***}} & \phantom{\textsuperscript{***}}59.1\phantom{\textsuperscript{***}} \\
    \cmidrule(lr){1-9}
    \addlinespace[0.4ex]
    \textit{No Concat}     & 2 & \phantom{\textsuperscript{***}}41.0\phantom{\textsuperscript{***}} & \phantom{\textsuperscript{***}}49.7\phantom{\textsuperscript{***}} & \phantom{\textsuperscript{***}}60.5\phantom{\textsuperscript{***}} & \phantom{\textsuperscript{***}}66.5\phantom{\textsuperscript{***}} & \phantom{\textsuperscript{***}}65.0\phantom{\textsuperscript{***}} & \phantom{\textsuperscript{***}}60.0\phantom{\textsuperscript{***}} & \phantom{\textsuperscript{***}}57.1\phantom{\textsuperscript{***}} \\
                         & 4 & \phantom{\textsuperscript{***}}46.5\phantom{\textsuperscript{***}} & \phantom{\textsuperscript{***}}53.6\textsuperscript{*}\phantom{\textsuperscript{**}} & \phantom{\textsuperscript{***}}59.5\phantom{\textsuperscript{***}} & \phantom{\textsuperscript{***}}68.6\phantom{\textsuperscript{***}} & \phantom{\textsuperscript{***}}63.5\phantom{\textsuperscript{***}} & \phantom{\textsuperscript{***}}69.5\phantom{\textsuperscript{***}} & \phantom{\textsuperscript{***}}60.2\phantom{\textsuperscript{***}} \\
    \addlinespace[0.8ex]
    \textit{Pair Example Concat} & 2 & \phantom{\textsuperscript{***}}43.0\phantom{\textsuperscript{***}} & \phantom{\textsuperscript{***}}51.4\phantom{\textsuperscript{***}} & \phantom{\textsuperscript{***}}62.5\phantom{\textsuperscript{***}} & \phantom{\textsuperscript{***}}63.5\phantom{\textsuperscript{***}} & \phantom{\textsuperscript{***}}66.0\phantom{\textsuperscript{***}} & \phantom{\textsuperscript{***}}56.5\phantom{\textsuperscript{***}} & \phantom{\textsuperscript{***}}57.2\phantom{\textsuperscript{***}} \\
                         & 4 & \phantom{\textsuperscript{***}}47.0\phantom{\textsuperscript{***}} & \phantom{\textsuperscript{***}}52.5\phantom{\textsuperscript{***}} & \phantom{\textsuperscript{***}}57.9\phantom{\textsuperscript{***}} & \phantom{\textsuperscript{***}}64.0\phantom{\textsuperscript{***}} & \phantom{\textsuperscript{***}}67.5\phantom{\textsuperscript{***}} & \phantom{\textsuperscript{***}}64.0\phantom{\textsuperscript{***}} & \phantom{\textsuperscript{***}}58.8\phantom{\textsuperscript{***}} \\
    \addlinespace[0.8ex]
    \textit{Examples Concat}    & 2 & \phantom{\textsuperscript{***}}47.2\phantom{\textsuperscript{***}} & \phantom{\textsuperscript{***}}50.3\phantom{\textsuperscript{***}} & \phantom{\textsuperscript{***}}59.0\phantom{\textsuperscript{***}} & \phantom{\textsuperscript{***}}64.0\phantom{\textsuperscript{***}} & \phantom{\textsuperscript{***}}68.0\phantom{\textsuperscript{***}} & \phantom{\textsuperscript{***}}62.5\phantom{\textsuperscript{***}} & \phantom{\textsuperscript{***}}58.5\phantom{\textsuperscript{***}} \\
                         & 4 & \phantom{\textsuperscript{***}}58.5\textsuperscript{**}\phantom{\textsuperscript{*}} & \phantom{\textsuperscript{***}}46.4\phantom{\textsuperscript{***}} & \phantom{\textsuperscript{***}}64.5\textsuperscript{*}\phantom{\textsuperscript{**}} & \phantom{\textsuperscript{***}}63.0\phantom{\textsuperscript{***}} & \phantom{\textsuperscript{***}}65.0\phantom{\textsuperscript{***}} & \phantom{\textsuperscript{***}}57.5\phantom{\textsuperscript{***}} & \phantom{\textsuperscript{***}}59.2\phantom{\textsuperscript{***}} \\
    \addlinespace[0.8ex]
    \textit{Test Concat} & \phantom{$^\dagger$}0$^\dagger$ & \phantom{\textsuperscript{***}}66.0\textsuperscript{***} & \phantom{\textsuperscript{***}}54.2\textsuperscript{*}\phantom{\textsuperscript{**}} & \phantom{\textsuperscript{***}}64.0\phantom{\textsuperscript{***}} & \phantom{\textsuperscript{***}}72.5\phantom{\textsuperscript{***}} & \phantom{\textsuperscript{***}}70.5\phantom{\textsuperscript{***}} & \phantom{\textsuperscript{***}}\textbf{65.5}\phantom{\textsuperscript{***}} & \phantom{\textsuperscript{***}}65.4\phantom{\textsuperscript{***}} \\
                         & 2 & \phantom{\textsuperscript{***}}58.0\textsuperscript{**}\phantom{\textsuperscript{*}} & \phantom{\textsuperscript{***}}51.4\phantom{\textsuperscript{***}} & \phantom{\textsuperscript{***}}63.0\phantom{\textsuperscript{***}} & \phantom{\textsuperscript{***}}62.5\phantom{\textsuperscript{***}} & \phantom{\textsuperscript{***}}67.0\phantom{\textsuperscript{***}} & \phantom{\textsuperscript{***}}58.5\phantom{\textsuperscript{***}} & \phantom{\textsuperscript{***}}60.1\phantom{\textsuperscript{***}} \\
                         & 4 & \phantom{\textsuperscript{***}}63.5\textsuperscript{***} & \phantom{\textsuperscript{***}}\textbf{63.7}\textsuperscript{***} & \phantom{\textsuperscript{***}}59.0\phantom{\textsuperscript{***}} & \phantom{\textsuperscript{***}}\underline{73.0}\phantom{\textsuperscript{***}} & \phantom{\textsuperscript{***}}66.5\phantom{\textsuperscript{***}} & \phantom{\textsuperscript{***}}55.0\phantom{\textsuperscript{***}} & \phantom{\textsuperscript{***}}63.5\phantom{\textsuperscript{***}} \\
    \addlinespace[0.8ex]
    \textit{Examples$\&$Test Concat}   & 2 & \phantom{\textsuperscript{***}}63.5\textsuperscript{***} & \phantom{\textsuperscript{***}}\underline{56.4}\textsuperscript{**}\phantom{\textsuperscript{*}} & \phantom{\textsuperscript{***}}\textbf{74.0}\textsuperscript{**}\phantom{\textsuperscript{*}} & \phantom{\textsuperscript{***}}67.0\phantom{\textsuperscript{***}} & \phantom{\textsuperscript{***}}70.5\phantom{\textsuperscript{***}} & \phantom{\textsuperscript{***}}62.5\phantom{\textsuperscript{***}} & \phantom{\textsuperscript{***}}\underline{65.7}\phantom{\textsuperscript{***}} \\
                         & 4 & \phantom{\textsuperscript{***}}\underline{66.5}\textsuperscript{***} & \phantom{\textsuperscript{***}}55.3\textsuperscript{**}\phantom{\textsuperscript{*}} & \phantom{\textsuperscript{***}}\underline{70.0}\textsuperscript{*}\phantom{\textsuperscript{**}} & \phantom{\textsuperscript{***}}71.0\phantom{\textsuperscript{***}} & \phantom{\textsuperscript{***}}\textbf{74.5}\phantom{\textsuperscript{***}} & \phantom{\textsuperscript{***}}64.0\phantom{\textsuperscript{***}} & \phantom{\textsuperscript{***}}\textbf{66.9}\phantom{\textsuperscript{***}} \\
                         & 6 & \phantom{\textsuperscript{***}}\underline{66.5}\textsuperscript{***} & \phantom{\textsuperscript{***}}50.6\phantom{\textsuperscript{***}} & \phantom{\textsuperscript{***}}66.0\textsuperscript{*}\phantom{\textsuperscript{**}}  & \phantom{\textsuperscript{***}}\textbf{73.5}\phantom{\textsuperscript{***}} & \phantom{\textsuperscript{***}}71.0\phantom{\textsuperscript{***}} & \phantom{\textsuperscript{***}}60.5\phantom{\textsuperscript{***}} & \phantom{\textsuperscript{***}}64.7\phantom{\textsuperscript{***}} \\
                         & 8 & \phantom{\textsuperscript{***}}\textbf{67.5}\textsuperscript{***} & \phantom{\textsuperscript{***}}51.4\phantom{\textsuperscript{***}} & \phantom{\textsuperscript{***}}64.5\textsuperscript{*}\phantom{\textsuperscript{**}} & \phantom{\textsuperscript{***}}71.0\phantom{\textsuperscript{***}} & \phantom{\textsuperscript{***}}70.0\phantom{\textsuperscript{***}} & \phantom{\textsuperscript{***}}58.0\phantom{\textsuperscript{***}} & \phantom{\textsuperscript{***}}63.7\phantom{\textsuperscript{***}} \\
    \midrule
    Human Performance    & N/A & \phantom{\textsuperscript{***}}69.5\phantom{\textsuperscript{***}} & \phantom{\textsuperscript{***}}77.8\phantom{\textsuperscript{***}} & \phantom{\textsuperscript{***}}83.0\phantom{\textsuperscript{***}} & \phantom{\textsuperscript{***}}71.0\phantom{\textsuperscript{***}} & \phantom{\textsuperscript{***}}78.0\phantom{\textsuperscript{***}} & \phantom{\textsuperscript{***}}77.5\phantom{\textsuperscript{***}} & \phantom{\textsuperscript{***}}76.2\phantom{\textsuperscript{***}} \\
    \bottomrule
  \end{tabular}
  \caption{Evaluation of AudioJudge design choices using GPT-4o-Audio on audio characteristic detection tasks. Accuracy (\%) is reported for pairwise comparisons. Bold = best; Underline = second best. Prn = pronunciation, Speed = speaking rate, SkID = speaker identification, SOM = SOMOS, TMH = TMHINTQ, ThaM = ThaiMOS. *, **, *** denote statistically significance over baseline at $p$ < 0.05, 0.01, 0.001, respectively. $^\dagger$This setting is also the 0-shot setting for \textit{Examples\&Test Concat}.}
  \label{tab:fine_grained_design_choices}
\end{table*}

\section{Audio Characteristic Detection}
\label{section:fine_grained}

\subsection{Task Definition and Motivation}
\label{subsec:fine_grained_motivation}

Audio characteristic detection focuses on evaluating specific, measurable properties of speech signals. Rather than training specialized models for each characteristic~\citep{desplanques2020ecapa, utmos_interspeech, mosanetplus_icme, wang2024enabling}, we explore whether AudioJudge can serve as a unified framework through prompting alone. We assess this from two key perspectives.

From the perspective of 
\textbf{Paralinguistic Feature Detection}, we look at whether LAM can accurately detect subtle speech variations by quantifying: 
\vspace{-0.2em}
\begin{itemize}[leftmargin=*]
\setlength\itemsep{-0.2em}
    \item \textbf{Pronunciation accuracy:} Do two pronunciations of the same word match?
    \item \textbf{Speaking rate detection:} Which speech utterance is spoken faster? 
    \item \textbf{Speaker identification:} Are these audios from the same speaker? 
\end{itemize}

From the aspect of \textbf{Speech Quality Assessment}, we examine whether LAM can distinguish speech clarity, intelligibility, and naturalness, using existing speech quality evaluation datasets, across different conditions and languages as follows:
\vspace{-0.2em}

\begin{itemize}[leftmargin=*]
\setlength\itemsep{-0.2em}
    \item \textbf{SOMOS:} Naturalness assessment of synthesized English speech;
    \item \textbf{TMHINTQ:} Mandarin Chinese speech quality under various noise conditions;  
    \item \textbf{ThaiMOS:} Pronunciation accuracy evaluation of synthesized Thai speech.
\end{itemize}
Each evaluation uses a pairwise comparison format where LAMs determine which of two audio samples better exhibits the target characteristic. Detailed dataset descriptions are in Appendix~\ref{appendix:fine-grained_data}.
\subsection{Results and Analysis}
\label{subsec:fine_grained_results}

\paragraph{Understanding Task Difficulty and Human Performance Ceiling.} 
Before interpreting model performance, we aim to understand the inherent difficulty of the tasks. To that end, we conducted an independent human evaluation using 2 annotators who were not involved in the original dataset annotations, with the same instructions as LAMs. Annotators achieved 69.5\%-83.0\% accuracy across tasks (as shown at the bottom of Table~\ref{tab:fine_grained_design_choices}), reflecting the subjective nature of the tasks, such as in the standards for speech quality and pronunciation. These human performance levels set realistic upper bounds for automated evaluation.

We then evaluate AudioJudge design choices introduced in Section~\ref{section:designing_lam} on audio characteristic detection tasks using GPT-4o-Audio. The experimental results in Table~\ref{tab:fine_grained_design_choices} reveal the following findings:

\noindent \textbf{1) Baseline fails on paralinguistic tasks.} With basic prompting, the accuracy on pronunciation (46.0\%) and speaking rate (46.9\%) approximates random guessing, showing that even strong LAMs struggle to pick up paralinguistic cues without guidance. In contrast, speech quality tasks show substantially better baseline performance.

\noindent \textbf{2) Transcript information provides minimal benefits.} While adding ground truth transcript improves pronunciation detection significantly (46.0\% $\rightarrow$ 63.0\%, $p<0.01$), other tasks show minimal improvement or even degrade, leading to lower average performance. As textual information is not helpful for most paralinguistic judgments, we do not investigate this approach further.

\noindent \textbf{3) ICL without audio concatenation yields minimal improvements.} Traditional in-context learning with separately presented audio examples yields only marginal gains over zero-shot. Even with 4-shot examples, most tasks show modest improvements, with only speaking rate achieving statistical significance (46.9\%$\rightarrow$53.6\%, $p<0.05$), indicating that ICL without audio concatenation is insufficient for complex audio evaluations.

\noindent \textbf{4) Audio concatenation strategies show substantial benefits.} Concatenating test audios (\textit{Test Concat}) alone produces meaningful improvements across multiple tasks. Compared to the baseline, 0-shot \textit{Test Concat} achieves significant gains on pronunciation (46.0\%$\rightarrow$66.0\%, $p<0.001$) and speaking rate (46.9\%$\rightarrow$54.2\%, $p<0.05$). This suggests that eliminating modality transitions between audio segments helps LAMs focus on direct audio comparison.

\noindent \textbf{5) Examples$\&$Test Concat emerges as the optimal strategy.} Using 4-shot \textit{Examples$\&$Test Concat}, we achieve the best average performance, with significant gains over the baseline in pronunciation (46.0\%$\rightarrow$66.5\%, $p<0.001$), speaking rate (46.9\%$\rightarrow$55.3\%, $p<0.01$), and speaker identification (61.5\%$\rightarrow$70.0\%, $p<0.05$). For speech quality tasks, the method also shows improvements and approaches human performance. However, LAMs still face challenges with certain paralinguistic tasks, particularly speaking rate detection where a large gap with human performance remains.

\noindent \textbf{6) Diminishing returns beyond 4-shot examples.} Given the superior performance of the \textit{Examples\&Test Concat} method within the 4-shot range, we extended our analysis to include 6 and 8 examples. However, this yielded minimal gains and occasionally decreased performance. The 4-shot setup appears to provide an optimal balance between providing sufficient guidance and avoiding information overload.

\paragraph{Take-aways} 
The current AudioJudge, with basic prompting, struggles to distinguish paralinguistic clues. Incorporating audio concatenation and ICL examples significantly improves performance, bringing it closer to human performance on tasks such as pronunciation. \textbf{\textit{Examples\&Test Concatenation} with 4-shot examples emerges as the optimal configuration}, which we adopt for subsequent experiments in Section~\ref{section:system_level}. However, these prompting engineering techniques remain insufficient for certain aspects like speaking rate detection.

\subsection{Comparison with Specialized Models}
\label{ssec:comparison_trained}
To contextualize AudioJudge performance, we also compare it against existing specialized neural networks trained specifically for speech quality assessment. Table~\ref{tab:audio_judge_vs_baselines} presents results for UTMOS~\citep{utmos_interspeech}, MOSANET+~\citep{mosanetplus_icme}, and SALMONN-FT~\citep{wang2024enabling}—models fine-tuned on MOS-labeled data.
\begin{table}[t]
  \centering
  \small         
  \tabcolsep=2.0mm
  \begin{tabular}{l c c c c}
    \toprule
    \textbf{Method} & \textbf{Data Req} & {SOM} & {TMH} & {ThaM}\\
    \midrule
    UTMOS                    & 100 hrs$^\dagger$ & 77.5 & 71.5 & 53.5 \\
    MOSANET+                 & 25 hrs$^\dagger$& \textbf{85.0} & \textbf{77.5} & 62.5 \\
    SALMONN-FT               & 650 hrs$^\dagger$ & 82.0 & 61.0 & 58.5 \\
    \midrule
    AudioJudge                & \textbf{<0.2 hrs}& 71.0 & 74.5 & \textbf{64.0} \\
    \bottomrule
  \end{tabular}
    \caption{AudioJudge vs. specialized baselines on speech quality assessment. Data Req = estimated human annotation time for data used for training/in-context learning. $^\dagger$We estimate by \#annotations times 6 seconds as the average time for each annotation. AudioJudge uses \textit{Examples$\&$Test Concat} 4-shot.}
  \label{tab:audio_judge_vs_baselines}
\end{table}

The results show that specialized networks achieve superior performance on in-domain tasks (SOMOS and TMHINTQ), reflecting the benefits of task-specific training with extensive labeled data. However, their performance degrades substantially on out-of-domain dataset such as ThaiMOS, where they underperform despite their training overhead. In contrast, AudioJudge demonstrates more consistent cross-domain performance with minimal data requirements through in-context learning, achieving competitive results on ThaiMOS (64.0\%) and even outperforming some specialized models on cross-domain tasks. \textbf{This highlights LAMs as an effective alternative for speech quality assessment}, especially for diverse languages or low-resource evaluation settings.

\label{subsec:system_level_results}
\begin{table*}[!t]
  \small
  \tabcolsep=2.5mm
  \centering
  \begin{tabular}{l c c c c}
    \toprule
    \multirow{2}{*}{\textbf{Method}} & \multicolumn{2}{c}{\textbf{ChatbotArena-Spoken (Lexical)}} & \multicolumn{2}{c}{\textbf{SpeakBench (Multi-Aspect)}} \\
    \cmidrule(lr){2-3} \cmidrule(lr){4-5}
     & \textbf{GPT-4o} & \textbf{Gemini-2.5} & \textbf{GPT-4o} & \textbf{Gemini-2.5} \\
    \midrule
    Random Guess         & 0.000 & 0.000 & 0.000 & 0.000 \\
    AudioJudge             & 0.902 & 0.805 & 0.731 & 0.846 \\
    \midrule
    AudioJudge + ICL   & \textbf{0.931} & \textbf{0.877} & 0.775 & 0.857 \\
    \midrule
    Multi-Aspect AudioJudge & - & - & 0.802 & \textbf{0.912} \\
    Multi-Aspect AudioJudge + ICL & - & - & \textbf{0.846} & 0.857 \\
    \bottomrule
  \end{tabular}
  \caption{Human preference simulation with GPT-4o-Audio and Gemini-2.5-Flash. Spearman correlations are reported for ChatbotArena-Spoken (lexical content) and SpeakBench (multi-aspect evaluation). ICL refers to the \textit{Examples$\&$Test Concat} with 4-shot examples. "-" indicates settings not applicable to the respective datasets.}
  \label{tab:preference_simulation_results}
\end{table*}

\section{Human Preference Correlation}
\label{section:system_level}

\subsection{Task Definition and Motivation}
\label{subsec:system_level_motivation}

Human preference correlation focuses on ranking systems, enabling automated benchmarking and system comparison. We develop two datasets targeting different aspects of system-level evaluation:

\textbf{Lexical Content Evaluation:} Can LAMs rank systems based on lexical content quality when delivered through speech? We evaluate this using \textbf{ChatbotArena-Spoken}, where we synthesize spoken versions of text conversations from a subset of ChatbotArena (filtered to be suitable for a conversation format). Since the original annotations assess lexical content quality, this tests whether LAMs can maintain ranking accuracy when the same content is presented auditorily. \footnote{We provide modality consistency analysis (e.g., text-to-text, or audio-to-audio) in Appendix~\ref{subsec:cross_modality}.}

\textbf{Multi-Aspect Speech Evaluation:} Can LAMs simulate human preferences that encompass lexical content, speech quality, and paralinguistic appropriateness? We test this using \textbf{SpeakBench}, a \textit{speech-in speech-out} evaluation dataset designed to assess whether a system can (1) understand a spoken instruction and (2) generate a spoken response that not only conveys appropriate content but also expresses the required paralinguistic feature such as pronunciation (accents, tones), speaking style/emotion, prosody \& delivery (volume, pitch, speed), or non-linguistic sound effects (whistling, animal sounds). SpeakBench comprises 82 instructions, and we collect 508 
human judgments across 13 speech-in speech-out systems.\footnote{Detailed dataset descriptions and human annotation procedures are provided in Appendix~\ref{appendix:system_level_data}}

\subsection{Results and Analysis}

Building on prior findings (in Section \ref{section:fine_grained}), we evaluate overall  human preference correlation using the best setup: 4-shot \textit{Examples$\&$Test Concatenation}. We assess two leading LAMs—GPT-4o-Audio and Gemini-2.5-Flash—reporting Spearman correlations between LAM judgments and human preferences. Also, the multi-aspect nature of SpeakBench enables evaluation of the multi-aspect ensemble method introduced in Section~\ref{ssec:multi_aspect_ensemble}.

Table~\ref{tab:preference_simulation_results} presents our system-level preference simulation results, highlighting several key findings:

\noindent \textbf{1) Strong baseline performance across both datasets.} Both LAMs demonstrate impressive zero-shot performance, with GPT-4o-Audio achieving 0.902 correlation on ChatbotArena-Spoken and 0.731 on SpeakBench, while Gemini-2.5-Flash reaches 0.805 and 0.846 respectively. This indicates that current LAMs are able to rank speech-in speech-out systems at a reliable level.

\noindent \textbf{2) Consistent improvements from audio concatenation and in-context learning.} The \textit{Examples$\&$Test Concat} 4-shot setup provides improvements over baseline performance in both models and datasets. On ChatbotArena-Spoken, GPT-4o-Audio improves from 0.902 to 0.931, while Gemini-2.5-Flash gains from 0.805 to 0.877. Similarly, SpeakBench shows improvements from 0.731 to 0.775 for GPT-4o-Audio and from 0.846 to 0.857 for Gemini-2.5-Flash.

\noindent \textbf{3) Multi-aspect ensemble shows superior performance.} For SpeakBench, the multi-aspect ensemble approach achieves higher correlations: 0.802 for GPT-4o-Audio and 0.912 for Gemini-2.5-Flash in the zero-shot setting. This represents a substantial improvement over single-judge approaches, demonstrating the value of specialized judges for different evaluation dimensions.

\noindent \textbf{4) Model-dependent effectiveness of combination strategies.} Interestingly, combining multi-aspect ensemble with \textit{Examples$\&$Test Concat} shows different effects across models. For GPT-4o-Audio, the combination further improves performance (0.802 $\rightarrow$ 0.846), while for Gemini-2.5-Flash, it slightly degrades performance (0.912 $\rightarrow$ 0.857). This suggests that optimal prompting strategies can be model-dependent and that the ensemble approach likely already saturates in performance.

\begin{figure}[t]
    \centering
        \centering
        \includegraphics[width=0.4\textwidth]{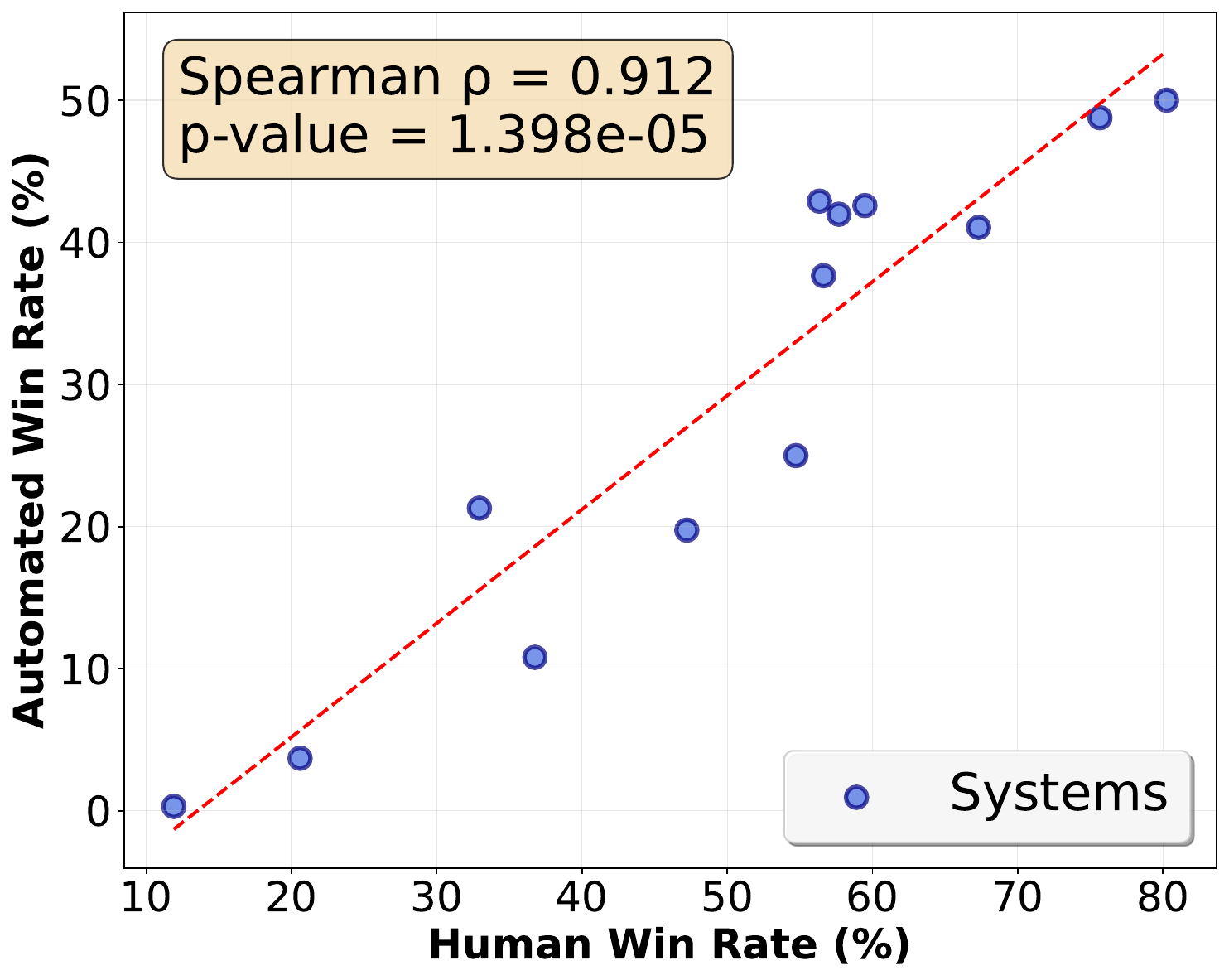}
    \caption{AudioJudge (Multi-Aspect Ensemble with Gemini-2.5-Flash) predictions and human preferences on SpeakBench. The analysis of this ranking is in Appendix~\ref{app:speakbench_ranking}, and a similar plot for ChatbotArena-Spoken is shown in Figure~\ref{fig:chatbot_arena_correlation} in Appendix~\ref{appendix:supplement}.\vspace{-0.25cm}}

    \label{fig:correlation_analysis}
\end{figure}

\paragraph{Takeaways.} With proper prompt engineering, AudioJudge achieves a strong correlation with human preferences, up to 0.93 on ChatbotArena-Spoken and 0.91 on SpeakBench (illustrated in Figure~\ref{fig:correlation_analysis}) -- a similar correlation level to AlpacaEval~\citep{alpaca_eval}, making it a potential solution to automated speech-based system benchmarking.

\section{Robustness Analysis of AudioJudge}
\label{sec:robustness}

This section assesses robustness and biases, critical for gauging AudioJudge's reliability. To isolate inherent robustness properties, all experiments use the zero-shot setup without audio concatenation.

\subsection{Robustness Against Noise}
\label{ssec:noise_robustness}

We test noise robustness by incrementally adding white Gaussian noise to ChatbotArena-Spoken audio samples, avoiding non-lexical tasks whose labels could shift under noise distortion.

\begin{figure}[t]
    \centering
    \includegraphics[width=0.91\linewidth, keepaspectratio]{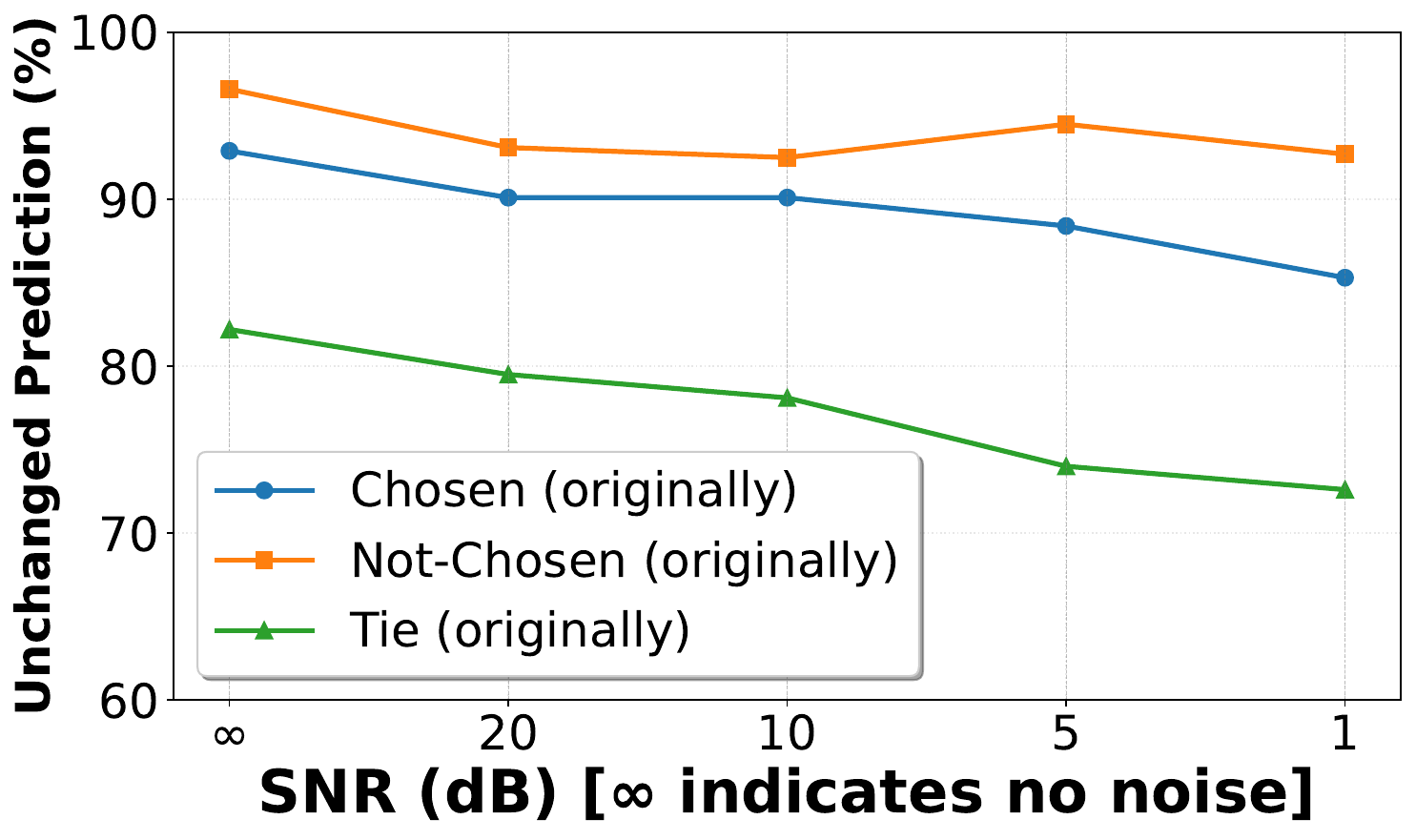}
    \caption{Noise robustness analysis. Percentage of unchanged GPT-4o-Audio predictions across varying Signal-to-Noise Ratios on ChatbotArena-Spoken.\vspace{-0.25cm}}
    \label{fig:snr_analysis}
\end{figure}
Figure~\ref{fig:snr_analysis} demonstrates that \textbf{GPT-4o-Audio maintains robust performance against noise perturbations.} Even at a low SNR of 1 dB, the unchanged prediction rates remain high across all prediction categories: 85\% for Chosen responses, 93\% for Not-Chosen responses, and 73\% for Tie decisions. These values significantly exceed the expected 33\% unchanged rate under complete audio corruption, indicating strong resistance to acoustic noise. This suggests that LAMs are likely optimized for content extraction under noise, as demonstrated by reliable performance even in noisy audio conditions.

\subsection{Verbosity Bias}
\label{ssec:verbosity_bias}

Humans and LLM judges exhibit verbosity bias—preferring longer responses when content is otherwise equal~\citep{saito2023verbosity}. We test whether LAMs have this bias by analyzing tie-rated examples. For non-lexical tasks, analyzing this bias is challenging due to confounding factors—longer responses may include words that are harder to pronounce or exhibit different prosodic patterns. Hence, we focus on ChatbotArena-Spoken.

\begin{figure}[t]
    \centering
    \includegraphics[width=0.92\linewidth, keepaspectratio]{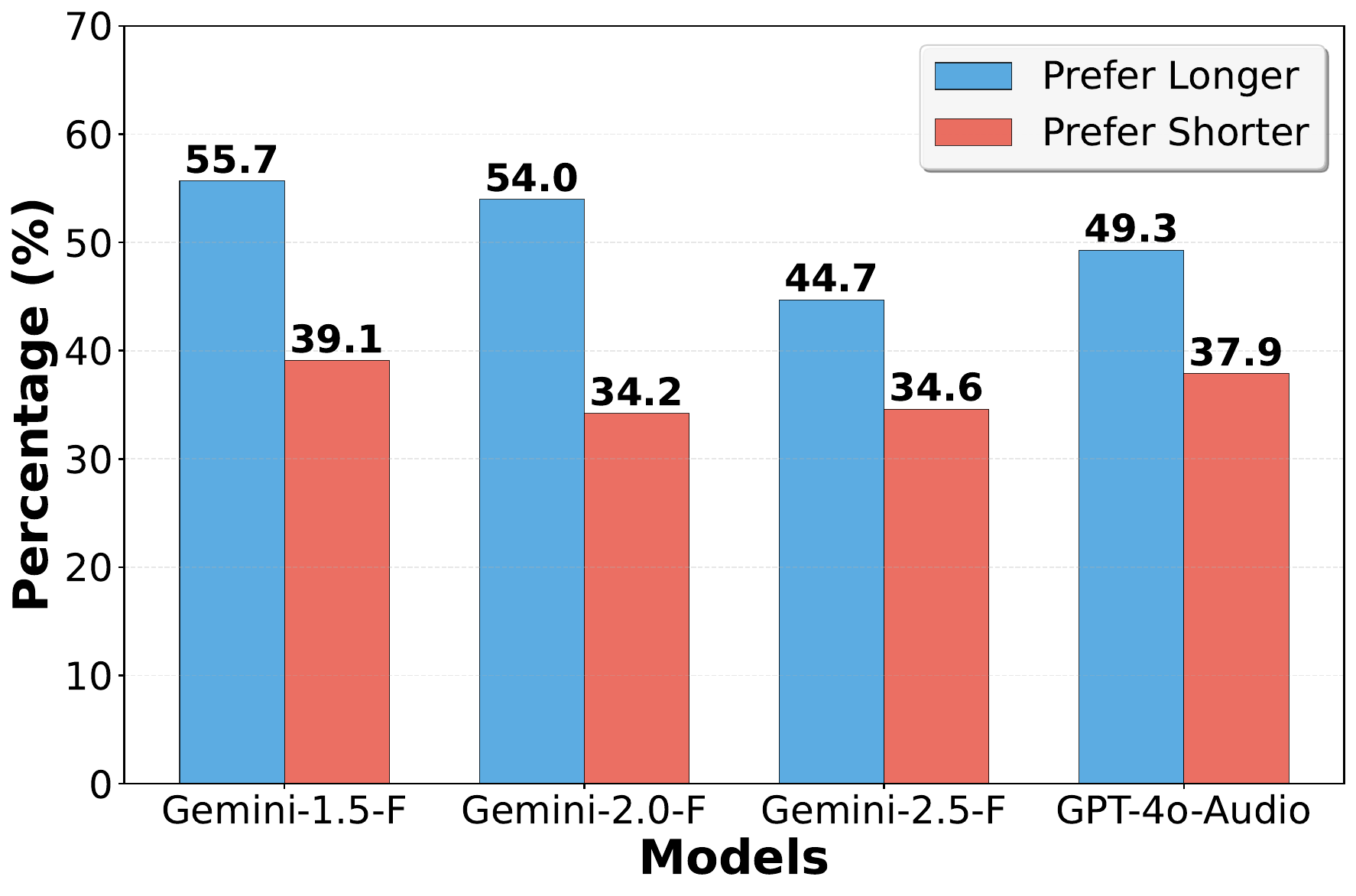}
    \caption{Verbosity bias on ChatbotArena‑Spoken. Percentage of judge preferences when two equally rated outputs differ in length--categorized as a tie, preference for longer output, or preference for shorter output.}
    \label{fig:verbosity_bias}
\end{figure}

Specifically, we examine LAM preferences on tie-rated \textit{audio} examples where one response’s transcript is at least 5 tokens longer than the other. As shown in Figure~\ref{fig:verbosity_bias}, \textbf{all models exhibit a preference for longer speech responses}. Bootstrap tests confirm that this bias toward longer responses is significant in all models ($p < 0.01$), reflecting systematic verbosity bias similar to prior text LLM-as-a-judge findings~\cite{saito2023verbosity, zheng2023judging}. Despite this, GPT-4o-Audio still achieves correlation $\rho$ > 0.9 because this bias primarily affects examples that are close in quality (e.g., tie-rated examples account for only only about 20\% of our data), and the verbosity bias does not favor specific models—meaning the relative ranking order remains preserved.

\subsection{Positional Bias}
\label{ssec:positional_bias}

Positional bias refers to systematic preferences for responses based on presentation order rather than quality. We measure it by presenting the same audio pair in both orders (A-B and B-A) and identifying cases where models consistently favor the first or second position, regardless of the content.

\begin{figure}[t]
    \centering
    \includegraphics[width=0.92\linewidth, keepaspectratio]{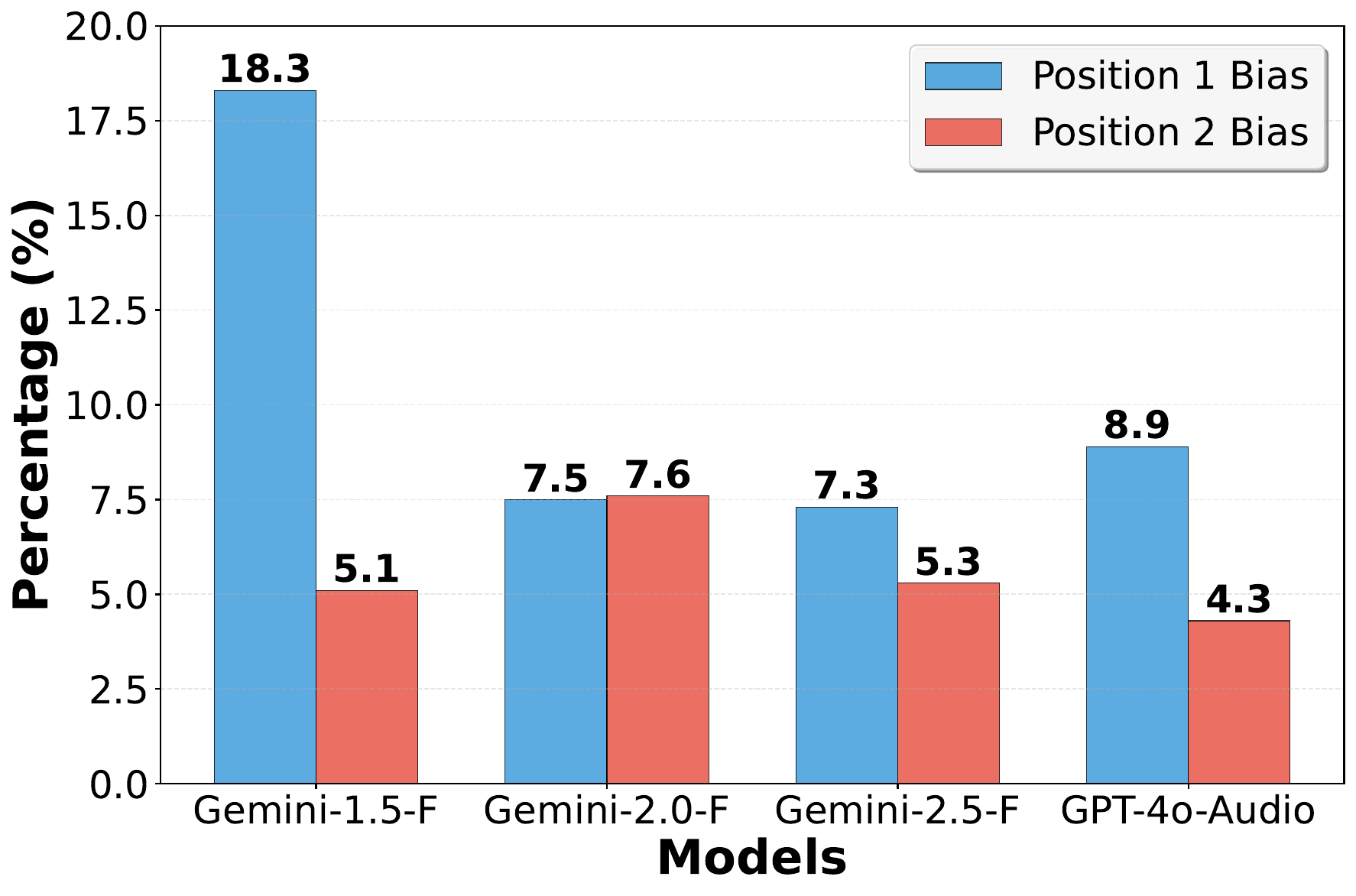}
    \caption{Positional bias on ChatbotArena-Spoken. Position 1 Bias and Position 2 Bias indicate the percentage of cases where the model consistently prefers the first or second response.\vspace{-0.25cm}}
    \label{fig:positional_bias_lexical}
\end{figure}

In lexical content evaluation, Figure~\ref{fig:positional_bias_lexical} shows that \textbf{GPT-4o-Audio, Gemini-1.5-Flash, and Gemini-2.5-Flash all favor the first position} (with bootstrap $p < 0.05$), whereas \textbf{Gemini-2.0-Flash displays no reliable directional bias }($p > 0.8$). Despite this bias, Gemini-2.5-Flash and GPT-4o-Audio remain stable on most datapoints (87.4\% and 89.8\% respectively). Gemini-1.5-Flash demonstrates both strong first-position bias (18.3\%) and lower overall stability (76.6\%), suggesting greater susceptibility to order effects.

\begin{figure}[t]
    \centering
    \includegraphics[width=0.92\linewidth, keepaspectratio]{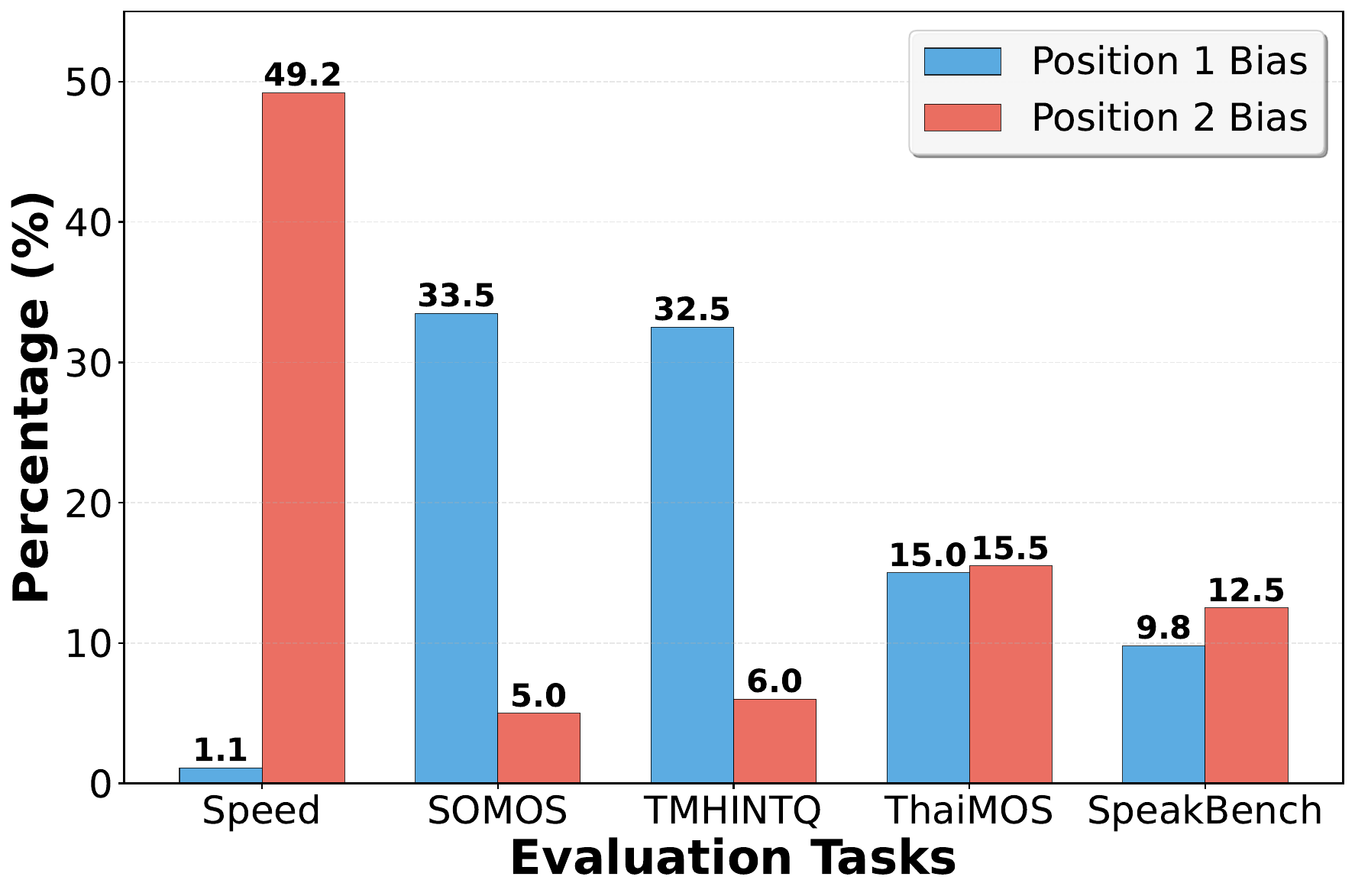}
   \caption{Positional bias on non-lexical datasets using GPT-4o-Audio. Position 1 and 2 Bias indicate the percentage of cases where the model consistently prefers the first or second response. Speed = speaking rate.\vspace{-0.25cm}}
    \label{fig:positional_bias_nonlexical}
\end{figure}

In non-lexical evaluation, positional effects are even more pronounced for speaking rate, speech quality, and SpeakBench as shown in Figure~\ref{fig:positional_bias_nonlexical}. Similar to lexical evaluation, \textbf{GPT-4o-Audio shows a first-position bias on SOMOS and TMHINTQ }($p < 0.01$). In contrast, \textbf{speaking rate evaluation exhibits a large recency bias} ($p < 0.01$) where the model predicts the second audio as faster in 49.2\% of cases even after the order is reversed. \textbf{ThaiMOS and SpeakBench display more balanced positional preferences}, but the percentage of stable predictions is still lower than 80.0\%. In Appendix~\ref{appendix:positional_bias_task_difficulty}, we show that positional bias grows as the MOS gap between SOMOS clips narrows (indicating harder discrimination). This highlights that current LAMs are more susceptible to positional bias as task difficulty increases.

\section{Conclusion}

This study finds that, with current LAMs, AudioJudge with basic prompting still struggles with non-lexical judgments, often performing near random chance. Prompting engineering techniques (ICL with audio concatenation) raise performance, yet even the best current setups remain \textit{insufficient} to evaluate all non-lexical scenarios at the example level. However, at the coarser system level, AudioJudge correlates with human preferences strongly ($\rho$>0.9), enabling reliable automated benchmarking. Robustness analysis shows that AudioJudge is strongly resistant to noise but exhibits persistent verbosity and positional biases, indicating the need for careful experimental design in evaluation.

\newpage
\section{Limitations}

Despite our prompt engineering efforts, LAM performance on paralinguistic tasks remains significantly worse than human annotators, with particularly large gaps in speaking rate detection (77.8\% vs 55.3\%) and speaker identification (83.0\% vs 70.0\%), indicating fundamental challenges in current LAMs' auditory discrimination capabilities. The \textit{Examples$\&$Test Concat} and multi-aspect ensemble approaches, while effective, impose substantial cost, limiting their practical scalability. Despite being the first speech-in speech-out evaluation dataset focusing on generated speech with paralinguistic features, SpeakBench is relatively small in scale, currently comprising 82 instructions and 13 systems (1,066 total datapoints), so it may not fully capture the diversity of real-world speech-in speech-out evaluation scenarios. Future work can look into extending this evaluation dataset in more speech-in speech-out scenarios as well as more models or systems.

\section*{Acknowledgements}
We appreciate the feedback provided by SALT members. We are thankful for computing support provided by the Stanford HAI-GCP Cloud Credit Grants and OpenAI. This work is funded in part by ONR Grant N000142412532, Sloan Foundation and NSF grant IIS-2247357. 

\bibliography{custom}

\begin{thebibliography}{48}
\providecommand{\natexlab}[1]{#1}

\bibitem[{Ao et~al.(2024)Ao, Wang, Tian, Chen, Zhang, Lu, Wang, Li, and Wu}]{ao2024sdeval}
Junyi Ao, Yuancheng Wang, Xiaohai Tian, Dekun Chen, Jun Zhang, Lu~Lu, Yuxuan Wang, Haizhou Li, and Zhizheng Wu. 2024.
\newblock \href {https://openreview.net/forum?id=PnjbvbblGv} {{SD}-eval: A benchmark dataset for spoken dialogue understanding beyond words}.
\newblock In \emph{The Thirty-eight Conference on Neural Information Processing Systems Datasets and Benchmarks Track}.

\bibitem[{Bastianelli et~al.(2020)Bastianelli, Vanzo, Swietojanski, and Rieser}]{bastianelli2020slurp}
Emanuele Bastianelli, Andrea Vanzo, Pawel Swietojanski, and Verena Rieser. 2020.
\newblock Slurp: A spoken language understanding resource package.
\newblock \emph{arXiv preprint arXiv:2011.13205}.

\bibitem[{Brown et~al.(2020)Brown, Mann, Ryder, Subbiah, Kaplan, Dhariwal, Neelakantan, Shyam, Sastry, Askell et~al.}]{brown2020language}
Tom Brown, Benjamin Mann, Nick Ryder, Melanie Subbiah, Jared~D Kaplan, Prafulla Dhariwal, Arvind Neelakantan, Pranav Shyam, Girish Sastry, Amanda Askell, et~al. 2020.
\newblock Language models are few-shot learners.
\newblock \emph{Advances in neural information processing systems}, 33:1877--1901.

\bibitem[{Chen et~al.(2025)Chen, Hu, Wang, Wang, Chen, Zhang, Yang, and Chng}]{chen2025audio}
Chen Chen, Yuchen Hu, Siyin Wang, Helin Wang, Zhehuai Chen, Chao Zhang, Chao-Han~Huck Yang, and EngSiong Chng. 2025.
\newblock \href {https://openreview.net/forum?id=U42TkrEDzb} {Audio large language models can be descriptive speech quality evaluators}.
\newblock In \emph{The Thirteenth International Conference on Learning Representations}.

\bibitem[{Chen et~al.(2024{\natexlab{a}})Chen, Chen, Zhang, Wang, Liu, Zhou, Zhang, Wan, Zhou, and Sun}]{chen2024unified}
Dongping Chen, Ruoxi Chen, Shilin Zhang, Yaochen Wang, Yinuo Liu, Huichi Zhou, Qihui Zhang, Yao Wan, Pan Zhou, and Lichao Sun. 2024{\natexlab{a}}.
\newblock Mllm-as-a-judge: Assessing multimodal llm-as-a-judge with vision-language benchmark.
\newblock In \emph{Forty-first International Conference on Machine Learning}.

\bibitem[{Chen et~al.(2024{\natexlab{b}})Chen, Yue, Zhang, Gao, Tan, and Li}]{chen2024voicebench}
Yiming Chen, Xianghu Yue, Chen Zhang, Xiaoxue Gao, Robby~T Tan, and Haizhou Li. 2024{\natexlab{b}}.
\newblock Voicebench: Benchmarking llm-based voice assistants.
\newblock \emph{arXiv preprint arXiv:2410.17196}.

\bibitem[{Chen and Tsao(2022)}]{tmhint_interspeech}
Yu-Wen Chen and Yu~Tsao. 2022.
\newblock \href {https://doi.org/10.21437/Interspeech.2022-10153} {Inqss: a speech intelligibility and quality assessment model using a multi-task learning network}.
\newblock In \emph{Interspeech 2022}, pages 3088--3092.

\bibitem[{Chu et~al.(2024)Chu, Xu, Yang, Wei, Wei, Guo, Leng, Lv, He, Lin, Zhou, and Zhou}]{qwen2audio}
Yunfei Chu, Jin Xu, Qian Yang, Haojie Wei, Xipin Wei, Zhifang Guo, Yichong Leng, Yuanjun Lv, Jinzheng He, Junyang Lin, Chang Zhou, and Jingren Zhou. 2024.
\newblock Qwen2-audio technical report.
\newblock \emph{arXiv preprint arXiv:2407.10759}.

\bibitem[{de~Seyssel et~al.(2024)de~Seyssel, D{'}Avirro, Williams, and Dupoux}]{deseyssel2023emphassess}
Maureen de~Seyssel, Antony D{'}Avirro, Adina Williams, and Emmanuel Dupoux. 2024.
\newblock \href {https://doi.org/10.18653/v1/2024.emnlp-main.30} {{E}mph{A}ssess : a prosodic benchmark on assessing emphasis transfer in speech-to-speech models}.
\newblock In \emph{Proceedings of the 2024 Conference on Empirical Methods in Natural Language Processing}, pages 495--507, Miami, Florida, USA. Association for Computational Linguistics.

\bibitem[{D{\'e}fossez et~al.(2024)D{\'e}fossez, Mazar{\'e}, Orsini, Royer, P{\'e}rez, J{\'e}gou, Grave, and Zeghidour}]{defossez2024moshi}
Alexandre D{\'e}fossez, Laurent Mazar{\'e}, Manu Orsini, Am{\'e}lie Royer, Patrick P{\'e}rez, Herv{\'e} J{\'e}gou, Edouard Grave, and Neil Zeghidour. 2024.
\newblock Moshi: a speech-text foundation model for real-time dialogue.
\newblock \emph{arXiv preprint arXiv:2410.00037}.

\bibitem[{Deshmukh et~al.(2024)Deshmukh, Alharthi, Elizalde, Gamper, Ismail, Singh, Raj, and Wang}]{deshmukh2024pam}
Soham Deshmukh, Dareen Alharthi, Benjamin Elizalde, Hannes Gamper, Mahmoud~Al Ismail, Rita Singh, Bhiksha Raj, and Huaming Wang. 2024.
\newblock Pam: Prompting audio-language models for audio quality assessment.
\newblock \emph{arXiv preprint arXiv:2402.00282}.

\bibitem[{Desplanques et~al.(2020)Desplanques, Thienpondt, and Demuynck}]{desplanques2020ecapa}
Brecht Desplanques, Jenthe Thienpondt, and Kris Demuynck. 2020.
\newblock Ecapa-tdnn: Emphasized channel attention, propagation and aggregation in tdnn based speaker verification.
\newblock \emph{arXiv preprint arXiv:2005.07143}.

\bibitem[{Dong et~al.(2022)Dong, Li, Dai, Zheng, Ma, Li, Xia, Xu, Wu, Liu et~al.}]{dong2022survey}
Qingxiu Dong, Lei Li, Damai Dai, Ce~Zheng, Jingyuan Ma, Rui Li, Heming Xia, Jingjing Xu, Zhiyong Wu, Tianyu Liu, et~al. 2022.
\newblock A survey on in-context learning.
\newblock \emph{arXiv preprint arXiv:2301.00234}.

\bibitem[{Dubois et~al.(2023)Dubois, Li, Taori, Zhang, Gulrajani, Ba, Guestrin, Liang, and Hashimoto}]{dubois2023alpacafarm}
Yann Dubois, Xuechen Li, Rohan Taori, Tianyi Zhang, Ishaan Gulrajani, Jimmy Ba, Carlos Guestrin, Percy Liang, and Tatsunori Hashimoto. 2023.
\newblock \href {https://openreview.net/forum?id=4hturzLcKX} {Alpacafarm: A simulation framework for methods that learn from human feedback}.
\newblock In \emph{Thirty-seventh Conference on Neural Information Processing Systems}.

\bibitem[{Efstathiadis et~al.(2025)Efstathiadis, Yadav, and Abbas}]{efstathiadis2025llm}
Georgios Efstathiadis, Vijay Yadav, and Anzar Abbas. 2025.
\newblock Llm-based speaker diarization correction: A generalizable approach.
\newblock \emph{Speech Communication}, 170:103224.

\bibitem[{Fang et~al.(2024)Fang, Guo, Zhou, Ma, Zhang, and Feng}]{fang2024llama}
Qingkai Fang, Shoutao Guo, Yan Zhou, Zhengrui Ma, Shaolei Zhang, and Yang Feng. 2024.
\newblock Llama-omni: Seamless speech interaction with large language models.
\newblock \emph{arXiv preprint arXiv:2409.06666}.

\bibitem[{Held et~al.(2024)Held, Li, Ryan, Shi, Zhang, and Yang}]{held2024distilling}
William Held, Ella Li, Michael Ryan, Weiyan Shi, Yanzhe Zhang, and Diyi Yang. 2024.
\newblock Distilling an end-to-end voice assistant without instruction training data.
\newblock \emph{arXiv preprint arXiv:2410.02678}.

\bibitem[{Jiang et~al.(2025)Jiang, Lin, Bu, Du, Wang, and Li}]{jiang2025s2s}
Feng Jiang, Zhiyu Lin, Fan Bu, Yuhao Du, Benyou Wang, and Haizhou Li. 2025.
\newblock S2s-arena, evaluating speech2speech protocols on instruction following with paralinguistic information.
\newblock \emph{arXiv preprint arXiv:2503.05085}.

\bibitem[{Koizumi et~al.(2023)Koizumi, Zen, Karita, Ding, Yatabe, Morioka, Bacchiani, Zhang, Han, and Bapna}]{koizumi23_interspeech}
Yuma Koizumi, Heiga Zen, Shigeki Karita, Yifan Ding, Kohei Yatabe, Nobuyuki Morioka, Michiel Bacchiani, Yu~Zhang, Wei Han, and Ankur Bapna. 2023.
\newblock \href {https://doi.org/10.21437/Interspeech.2023-1584} {Libritts-r: A restored multi-speaker text-to-speech corpus}.
\newblock In \emph{Interspeech 2023}, pages 5496--5500.

\bibitem[{Korzekwa et~al.(2021)Korzekwa, Lorenzo-Trueba, Drugman, Calamaro, and Kostek}]{korzekwa2021weakly}
Daniel Korzekwa, Jaime Lorenzo-Trueba, Thomas Drugman, Shira Calamaro, and Bozena Kostek. 2021.
\newblock Weakly-supervised word-level pronunciation error detection in non-native english speech.
\newblock \emph{arXiv preprint arXiv:2106.03494}.

\bibitem[{Lacombe et~al.(2024)Lacombe, Srivastav, and Gandhi}]{lacombe-etal-2024-parler-tts}
Yoach Lacombe, Vaibhav Srivastav, and Sanchit Gandhi. 2024.
\newblock Parler-tts.
\newblock \url{https://github.com/huggingface/parler-tts}.

\bibitem[{Latif et~al.(2023)Latif, Usama, Malik, and Schuller}]{latif2023emotionchatgpt}
Siddique Latif, Muhammad Usama, Mohammad~Ibrahim Malik, and Bj{\"o}rn~W Schuller. 2023.
\newblock Can large language models aid in annotating speech emotional data? uncovering new frontiers.
\newblock \emph{arXiv preprint arXiv:2307.06090}.

\bibitem[{Li et~al.(2025)Li, Held, Ryan, Pipatanakul, Manakul, Zhu, and Yang}]{li2025mind}
Minzhi Li, William~Barr Held, Michael~J Ryan, Kunat Pipatanakul, Potsawee Manakul, Hao Zhu, and Diyi Yang. 2025.
\newblock Mind the gap! static and interactive evaluations of large audio models.
\newblock \emph{arXiv preprint arXiv:2502.15919}.

\bibitem[{Li et~al.(2023)Li, Zhang, Dubois, Taori, Gulrajani, Guestrin, Liang, and Hashimoto}]{alpaca_eval}
Xuechen Li, Tianyi Zhang, Yann Dubois, Rohan Taori, Ishaan Gulrajani, Carlos Guestrin, Percy Liang, and Tatsunori~B. Hashimoto. 2023.
\newblock Alpacaeval: An automatic evaluator of instruction-following models.
\newblock \url{https://github.com/tatsu-lab/alpaca_eval}.

\bibitem[{Liusie et~al.(2024)Liusie, Manakul, and Gales}]{liusie-etal-2024-llm}
Adian Liusie, Potsawee Manakul, and Mark Gales. 2024.
\newblock \href {https://aclanthology.org/2024.eacl-long.8/} {{LLM} comparative assessment: Zero-shot {NLG} evaluation through pairwise comparisons using large language models}.
\newblock In \emph{Proceedings of the 18th Conference of the European Chapter of the Association for Computational Linguistics (Volume 1: Long Papers)}, pages 139--151, St. Julian{'}s, Malta. Association for Computational Linguistics.

\bibitem[{Lo et~al.(2019)Lo, Fu, Huang, Wang, Yamagishi, Tsao, and Wang}]{lo2019mosnet}
Chen-Chou Lo, Szu-Wei Fu, Wen-Chin Huang, Xin Wang, Junichi Yamagishi, Yu~Tsao, and Hsin-Min Wang. 2019.
\newblock Mosnet: Deep learning based objective assessment for voice conversion.
\newblock \emph{arXiv preprint arXiv:1904.08352}.

\bibitem[{Manakul et~al.(2024)Manakul, Sun, Sirichotedumrong, Tharnpipitchai, and Pipatanakul}]{manakul2024enhancing}
Potsawee Manakul, Guangzhi Sun, Warit Sirichotedumrong, Kasima Tharnpipitchai, and Kunat Pipatanakul. 2024.
\newblock Enhancing low-resource language and instruction following capabilities of audio language models.
\newblock \emph{arXiv preprint arXiv:2409.10999}.

\bibitem[{Maniati et~al.(2022)Maniati, Vioni, Ellinas, Nikitaras, Klapsas, Sung, Jho, Chalamandaris, and Tsiakoulis}]{somos_interspeech}
Georgia Maniati, Alexandra Vioni, Nikolaos Ellinas, Karolos Nikitaras, Konstantinos Klapsas, June~Sig Sung, Gunu Jho, Aimilios Chalamandaris, and Pirros Tsiakoulis. 2022.
\newblock \href {https://doi.org/10.21437/Interspeech.2022-10922} {Somos: The samsung open mos dataset for the evaluation of neural text-to-speech synthesis}.
\newblock In \emph{Interspeech 2022}, pages 2388--2392.

\bibitem[{Mittag et~al.(2021)Mittag, Naderi, Chehadi, and M{\"o}ller}]{mittag2021nisqa}
Gabriel Mittag, Babak Naderi, Assmaa Chehadi, and Sebastian M{\"o}ller. 2021.
\newblock Nisqa: A deep cnn-self-attention model for multidimensional speech quality prediction with crowdsourced datasets.
\newblock \emph{arXiv preprint arXiv:2104.09494}.

\bibitem[{{OpenAI}(2024)}]{openai2024gpt4otranscribe}
{OpenAI}. 2024.
\newblock Gpt-4o transcribe.
\newblock \url{https://platform.openai.com/docs/models/gpt-4o-transcribe}.
\newblock Accessed: 2024-12-17.

\bibitem[{Radford et~al.(2023)Radford, Kim, Xu, Brockman, McLeavey, and Sutskever}]{radford2023robust}
Alec Radford, Jong~Wook Kim, Tao Xu, Greg Brockman, Christine McLeavey, and Ilya Sutskever. 2023.
\newblock Robust speech recognition via large-scale weak supervision.
\newblock In \emph{International conference on machine learning}, pages 28492--28518. PMLR.

\bibitem[{Reddy et~al.(2021)Reddy, Gopal, and Cutler}]{reddy2021dnsmos}
Chandan~KA Reddy, Vishak Gopal, and Ross Cutler. 2021.
\newblock Dnsmos: A non-intrusive perceptual objective speech quality metric to evaluate noise suppressors.
\newblock In \emph{ICASSP 2021-2021 IEEE International Conference on Acoustics, Speech and Signal Processing (ICASSP)}, pages 6493--6497. IEEE.

\bibitem[{Saeki et~al.(2022)Saeki, Xin, Nakata, Koriyama, Takamichi, and Saruwatari}]{utmos_interspeech}
Takaaki Saeki, Detai Xin, Wataru Nakata, Tomoki Koriyama, Shinnosuke Takamichi, and Hiroshi Saruwatari. 2022.
\newblock \href {https://doi.org/10.21437/Interspeech.2022-439} {Utmos: Utokyo-sarulab system for voicemos challenge 2022}.
\newblock In \emph{Interspeech 2022}, pages 4521--4525.

\bibitem[{Saito et~al.(2023)Saito, Wachi, Wataoka, and Akimoto}]{saito2023verbosity}
Keita Saito, Akifumi Wachi, Koki Wataoka, and Youhei Akimoto. 2023.
\newblock Verbosity bias in preference labeling by large language models.
\newblock \emph{arXiv preprint arXiv:2310.10076}.

\bibitem[{Sakshi et~al.(2024)Sakshi, Tyagi, Kumar, Seth, Selvakumar, Nieto, Duraiswami, Ghosh, and Manocha}]{sakshi2024mmaumassivemultitaskaudio}
S~Sakshi, Utkarsh Tyagi, Sonal Kumar, Ashish Seth, Ramaneswaran Selvakumar, Oriol Nieto, Ramani Duraiswami, Sreyan Ghosh, and Dinesh Manocha. 2024.
\newblock \href {https://arxiv.org/abs/2410.19168} {Mmau: A massive multi-task audio understanding and reasoning benchmark}.
\newblock \emph{Preprint}, arXiv:2410.19168.

\bibitem[{Seki et~al.(2023)Seki, Takamichi, Saeki, and Saruwatari}]{seki_icassp2023}
Kentaro Seki, Shinnosuke Takamichi, Takaaki Saeki, and Hiroshi Saruwatari. 2023.
\newblock \href {https://doi.org/10.1109/ICASSP49357.2023.10095161} {Text-to-speech synthesis from dark data with evaluation-in-the-loop data selection}.
\newblock In \emph{ICASSP 2023 - 2023 IEEE International Conference on Acoustics, Speech and Signal Processing (ICASSP)}, pages 1--5.

\bibitem[{Shi et~al.(2025)Shi, Shim, Tian, Arora, Wu, Petermann, Yip, Zhang, Tang, Zhang, Alharthi, Huang, Saito, Han, Zhao, Donahue, and Watanabe}]{shi-etal-2025-versa}
Jiatong Shi, Hye-jin Shim, Jinchuan Tian, Siddhant Arora, Haibin Wu, Darius Petermann, Jia~Qi Yip, You Zhang, Yuxun Tang, Wangyou Zhang, Dareen~Safar Alharthi, Yichen Huang, Koichi Saito, Jionghao Han, Yiwen Zhao, Chris Donahue, and Shinji Watanabe. 2025.
\newblock \href {https://doi.org/10.18653/v1/2025.naacl-demo.19} {{VERSA}: A versatile evaluation toolkit for speech, audio, and music}.
\newblock In \emph{Proceedings of the 2025 Conference of the Nations of the Americas Chapter of the Association for Computational Linguistics: Human Language Technologies (System Demonstrations)}, pages 191--209, Albuquerque, New Mexico. Association for Computational Linguistics.

\bibitem[{Taal et~al.(2011)Taal, Hendriks, Heusdens, and Jensen}]{taal2011algorithm}
Cees~H Taal, Richard~C Hendriks, Richard Heusdens, and Jesper Jensen. 2011.
\newblock An algorithm for intelligibility prediction of time--frequency weighted noisy speech.
\newblock \emph{IEEE Transactions on audio, speech, and language processing}, 19(7):2125--2136.

\bibitem[{Tang et~al.(2024)Tang, Yu, Sun, Chen, Tan, Li, Lu, MA, and Zhang}]{tang2024salmonn}
Changli Tang, Wenyi Yu, Guangzhi Sun, Xianzhao Chen, Tian Tan, Wei Li, Lu~Lu, Zejun MA, and Chao Zhang. 2024.
\newblock \href {https://openreview.net/forum?id=14rn7HpKVk} {{SALMONN}: Towards generic hearing abilities for large language models}.
\newblock In \emph{The Twelfth International Conference on Learning Representations}.

\bibitem[{Wang et~al.(2025{\natexlab{a}})Wang, Zou, Lin, Sun, Liu, Zhang, Liu, Aw, and Chen}]{wang2024audiobench}
Bin Wang, Xunlong Zou, Geyu Lin, Shuo Sun, Zhuohan Liu, Wenyu Zhang, Zhengyuan Liu, AiTi Aw, and Nancy~F Chen. 2025{\natexlab{a}}.
\newblock Audiobench: A universal benchmark for audio large language models.
\newblock \emph{NAACL}.

\bibitem[{Wang et~al.(2025{\natexlab{b}})Wang, Yu, Yang, Tang, Li, Zhuang, Chen, Tian, Zhang, Sun et~al.}]{wang2024enabling}
Siyin Wang, Wenyi Yu, Yudong Yang, Changli Tang, Yixuan Li, Jimin Zhuang, Xianzhao Chen, Xiaohai Tian, Jun Zhang, Guangzhi Sun, et~al. 2025{\natexlab{b}}.
\newblock Enabling auditory large language models for automatic speech quality evaluation.
\newblock In \emph{ICASSP 2025-2025 IEEE International Conference on Acoustics, Speech and Signal Processing (ICASSP)}, pages 1--5. IEEE.

\bibitem[{Xiong et~al.(2024)Xiong, Wang, Guo, Ye, Fan, Gu, Huang, and Li}]{xiong2024llavacritic}
Tianyi Xiong, Xiyao Wang, Dong Guo, Qinghao Ye, Haoqi Fan, Quanquan Gu, Heng Huang, and Chunyuan Li. 2024.
\newblock Llava-critic: Learning to evaluate multimodal models.
\newblock \emph{arXiv preprint arXiv:2410.02712}.

\bibitem[{Yang et~al.(2023)Yang, Gu, Liu, Ghosh, Bulyko, and Stolcke}]{yang2023generative}
Chao-Han~Huck Yang, Yile Gu, Yi-Chieh Liu, Shalini Ghosh, Ivan Bulyko, and Andreas Stolcke. 2023.
\newblock Generative speech recognition error correction with large language models and task-activating prompting.
\newblock In \emph{2023 IEEE Automatic Speech Recognition and Understanding Workshop (ASRU)}, pages 1--8. IEEE.

\bibitem[{Yang et~al.(2025)Yang, Yang, Chen, Ma, Chen, Wang, Wang, Yang, Niu, Liu et~al.}]{yang2025emotts}
Guanrou Yang, Chen Yang, Qian Chen, Ziyang Ma, Wenxi Chen, Wen Wang, Tianrui Wang, Yifan Yang, Zhikang Niu, Wenrui Liu, et~al. 2025.
\newblock Emovoice: Llm-based emotional text-to-speech model with freestyle text prompting.
\newblock \emph{arXiv preprint arXiv:2504.12867}.

\bibitem[{Yang et~al.(2024)Yang, Xu, Liu, Chu, Jiang, Zhou, Leng, Lv, Zhao, Zhou et~al.}]{yang2024air}
Qian Yang, Jin Xu, Wenrui Liu, Yunfei Chu, Ziyue Jiang, Xiaohuan Zhou, Yichong Leng, Yuanjun Lv, Zhou Zhao, Chang Zhou, et~al. 2024.
\newblock Air-bench: Benchmarking large audio-language models via generative comprehension.
\newblock \emph{arXiv preprint arXiv:2402.07729}.

\bibitem[{Yang et~al.(2021)Yang, Chi, Chuang, Lai, Lakhotia, Lin, Liu, Shi, Chang, Lin et~al.}]{yang2021superb}
Shu-wen Yang, Po-Han Chi, Yung-Sung Chuang, Cheng-I~Jeff Lai, Kushal Lakhotia, Yist~Y Lin, Andy~T Liu, Jiatong Shi, Xuankai Chang, Guan-Ting Lin, et~al. 2021.
\newblock Superb: Speech processing universal performance benchmark.
\newblock \emph{arXiv preprint arXiv:2105.01051}.

\bibitem[{Zezario et~al.(2024)Zezario, Chen, Fu, Tsao, Wang, and Fuh}]{mosanetplus_icme}
Ryandhimas~E. Zezario, Yu-Wen Chen, Szu-Wei Fu, Yu~Tsao, Hsin-Min Wang, and Chiou-Shann Fuh. 2024.
\newblock \href {https://doi.org/10.1109/ICME57554.2024.10688047} {A study on incorporating whisper for robust speech assessment}.
\newblock In \emph{2024 IEEE International Conference on Multimedia and Expo (ICME)}, pages 1--6.

\bibitem[{Zheng et~al.(2023)Zheng, Chiang, Sheng, Zhuang, Wu, Zhuang, Lin, Li, Li, Xing et~al.}]{zheng2023judging}
Lianmin Zheng, Wei-Lin Chiang, Ying Sheng, Siyuan Zhuang, Zhanghao Wu, Yonghao Zhuang, Zi~Lin, Zhuohan Li, Dacheng Li, Eric Xing, et~al. 2023.
\newblock Judging llm-as-a-judge with mt-bench and chatbot arena.
\newblock \emph{Advances in Neural Information Processing Systems}, 36:46595--46623.

\end{thebibliography}
\newpage
\appendix
\section{Datasets}

\subsection{Audio Characteristic
Detection Datasets}
\label{appendix:fine-grained_data}
\subsubsection{Pronunciation Dataset}
This dataset consists of pairs of a Wiktionary reference recording with the same word spoken by GPT-4o-Audio. Two annotators, who are native English speakers, labelled whether the pronunciations of the two recordings match or not, resulting in binary labels. In total, this work makes use of 200 such pairs for pronunciation assessment. 

\subsubsection{Speaking Rate Dataset} 
This dataset also draws utterance pairs from LibriTTS-R, but each pair comes from a single speaker. The speaker rate is computed as phonemes (following the data preparation recipe of ParlerTTS \citep{lacombe-etal-2024-parler-tts}) divided by the utterance duration. The label indicates which utterance is faster. This work makes use of 187 such data points.

\subsubsection{Speaker Identification Dataset}
This dataset is built from LibriTTS-R \citep{koizumi23_interspeech} by sampling utterance pairs that either share the same speaker or come from different speakers, yielding a binary same/different task. We did not separate between different genders of speakers. This work makes use of 200 such data points.

\subsubsection{SOMOS Dataset}
The dataset is derived from the original SOMOS dataset~\citep{somos_interspeech}. The speech is in English, containing synthezised speech with crowd-sourced MOS ratings on a 1-5 scale on the ``naturalness" of speech. The samples are taken from 200 TTS systems of 100 English sentences randomly selected from the LJ Speech scripts. Given that the task of naturalness annotation can be highly subjective, this pairwise dataset only contains pairs where the average difference in MOS ratings is greater than 1.0. Due to the high cost incurred in examining various prompt setups, we sample 200 random pairs (out of 593 all pairs satisfying the MOS difference) for evaluation.

\subsubsection{TMHINTQ Dataset}
The dataset is derived from the TMHINT-Q dataset \citep{tmhint_interspeech}. The speech is in Mandarin Chinese, and noise of different types and levels was added to the clean speech. Human annotators were asked to score each audio on its ``quality" aspect on a 1-5 scale. The full pairwise mapped dataset contains 6475 pairs, and for this work we sample 200 pairs for evaluation.

\subsubsection{ThaiMOS Dataset}
\label{appendix:thaimos}
We selected 50 sentences from the Thai subset of CommonVoice transcripts. Speech outputs were synthesized using 11 different TTS systems, producing 600 audio files (50 sentences $\times$ 12 systems, including the original CommonVoice recordings). The 11 TTS systems are PyThaiTTS, Azure TTS systems (Niwat, Achara, Premwadee), BOTNOI TTS systems (spk13, spk7, spk30), gTTS, macOS, Google Cloud Platform TTS, and Seamless.

The audio samples were evaluated by 16 human subjects employed by DataWow. Each subject listened to the utterances and provided ratings on using the following guideline with three aspects:
\vspace{-0.2em}
\begin{itemize}[leftmargin=*]
\setlength\itemsep{-0.2em}
    \item Sound Quality (Noise Level): Evaluates the presence of noise and distortions in the audio file.
    \item Rhythm: Assesses the naturalness of pauses between words and sentences.
    \item {Pronunciation}: Measures the accuracy of phonetic articulation for each word.
\end{itemize}
Each aspect was rated on a Likert scale from 1 to 5, where a higher score indicates better performance. For this work, we make use of \textbf{pronunciation} as the main quality score. Similar to SOMOS and TMHINTQ, we sample 200 pairs for evaluation.

\subsection{Human Preference Simulation Datasets}
\label{appendix:system_level_data}

\subsubsection{ChatbotArena-Spoken Dataset} We assume that some conversations are suitable for both written and spoken formats. Based on this assumption, we leverage the existing ChatbotArena dataset~\citep{zheng2023judging} to simulate speech-based conversations using the following steps:
\begin{itemize}[leftmargin=*]
\setlength\itemsep{-0.2em}
    \item \textit{Step 1}: Starting from the original ChatbotArena containing 33K conversations, we keep only two-turn conversations.
    
    \item \textit{Step 2}: We employ GPT-4o-mini to filter out non-spoken-like rows using a specialized filtering prompt.

    \item \textit{Step 3}: For each item (user question, response A, response B) in each row, we synthesize speech using one voice randomly selected from 12 high-quality voices (bella, nicole, sarah, kore, aoede, puck, michael, fenrir, emma, isabella, fable, george) in KokoroTTS v0.19. This yields 7.8K data points, from which we randomly select 1K for evaluation.
\end{itemize}

\subsubsection{SpeakBench Dataset} We curate SpeakBench comprising 82 paralinguistic-focused instructions across four categories specifically targeting speech-in speech-out system evaluation. We used GPT-4o to expand 20 seed prompts into 100 instructions, then removed duplicates and non-nuanced items. The instructions are synthesized into speech via KokoroTTS.

Table~\ref{tab:speakbench_categories} presents the four instruction categories with their descriptions, examples, and counts in our final corpus.

\begin{table*}[h]
\small
\centering
\begin{tabular}{p{2cm}p{5cm}p{6cm}c}
\toprule
\textbf{Category} & \textbf{Description} & \textbf{Example} & \textbf{Count} \\
\midrule
Pronunciation & Instructions emphasizing regional or language-specific accent differences, pronunciation nuances, or tones & \textit{Teach me an example of Chinese Mandarin tones using the word 'ma' in different tones. First, show me how you pronounce all tones in one go, then explain each one.} & 16 \\
\midrule
Speaking Style, Emotion, Tone & Instructions about telling a story/narrative sometimes with a focus on an emotion, tone, or speaking style & \textit{Tell a bedtime story about a robot using a whispering voice.} & 19 \\
\midrule
Prosody \& Delivery & Instructions focusing on variations in volume, pitch, and speed & \textit{Perform a countdown from 10 to 1, starting with a slow, deliberate pace and accelerating as you approach zero.} & 28 \\
\midrule
Non-Linguistic Sound Effects & Instructions requiring imitation of non-verbal sounds like whistling, animal calls or mimicking Morse code & \textit{Whistle a short tune and then smoothly transition to saying the phrase 'Good morning, have a great day!'} & 19 \\
\bottomrule
\end{tabular}
\caption{SpeakBench instruction categories with descriptions, examples, and counts.}
\label{tab:speakbench_categories}
\end{table*}

For each speech-in speech-out system, we prompted them with the audio instruction and collected their audio response for evaluation. There are 13 speech-in speech-out systems in the dataset, including end-to-end models and cascaded pipelines:

\begin{itemize}[leftmargin=*]
\setlength\itemsep{-0.2em}
\item \textbf{End-to-end (proprietary):} GPT-4o-Audio (\texttt{gpt-4o-audio-preview-2024-12-17}); Gemini-2.0-Flash (\texttt{gemini-2.0-flash-exp})
\item \textbf{End-to-end (open-source):} Typhoon2-Audio~\citep{manakul2024enhancing}, Llama-Omni~\citep{fang2024llama}, and Moshi~\citep{defossez2024moshi}
\item \textbf{Speech-in Text-out LAM:} DiVA~\citep{held2024distilling} and Qwen2-Audio~\citep{qwen2audio}
\item \textbf{Text LLM + TTS:} GPT-4o + TTS, Gemini-2.0-Flash + TTS, Llama3 + TTS  
\item \textbf{ASR + Text LLM + TTS:} GCP Speech-to-Text + Llama3 + KokoroTTS
\end{itemize}

\textbf{Human Annotation.} To validate AudioJudge for speech-in speech-out system ranking, we collect 508 human judgments on random model pairs covering every SpeakBench instruction, using 4 annotators. Each annotation involves an instruction and responses from two candidate models, and an annotator selects which response is better or declares a tie.

\section{Prompts}
\label{appendix:prompts}
For each dataset in Section~\ref{section:fine_grained} and Section~\ref{section:system_level}, there are specific system prompts and user messages as instructions, as presented in Table~\ref{tab:system_prompts} and Table~\ref{tab:user_messages}. 

\subsection{Audio–text Concatenation Prompt Strategies}
\label{appendix:concatenation_prompt}
\begin{figure}[!ht]
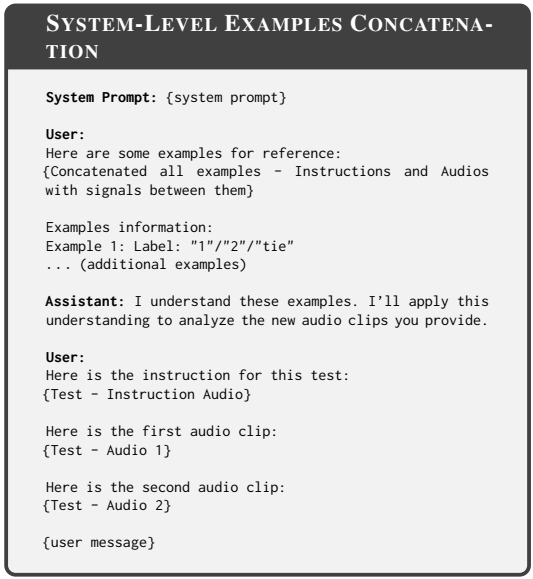

\centering
\tcbset{}
\begin{tcolorbox}[title=\textsc{System-Level Examples Concatenation},fonttitle=\small\bfseries,width=0.9\linewidth]
\tiny
\ttfamily
\textbf{System Prompt:} \{system prompt\}\\

\textbf{User:}\\
Here are some examples for reference:\\
\{Concatenated all examples - Instructions and Audios with signals between them\}\\

Examples information:\\
Example 1: Label: "1"/"2"/"tie"\\
$\ldots$ (additional examples)\\

\textbf{Assistant:} I understand these examples. I'll apply this understanding to analyze the new audio clips you provide.\\

\textbf{User:}\\
Here is the instruction for this test:\\
\{Test - Instruction Audio\}\\

Here is the first audio clip:\\
\{Test - Audio 1\}\\

Here is the second audio clip:\\
\{Test - Audio 2\}\\

\{user message\}
\end{tcolorbox}
\tcbset{reset}
\caption{System-level Examples Concatenation method: All example instructions and response audios are stitched into one long waveform, while test files remain separate.}
\label{fig:speakbench_aggregate_example}
\end{figure}

\begin{figure}[!ht]
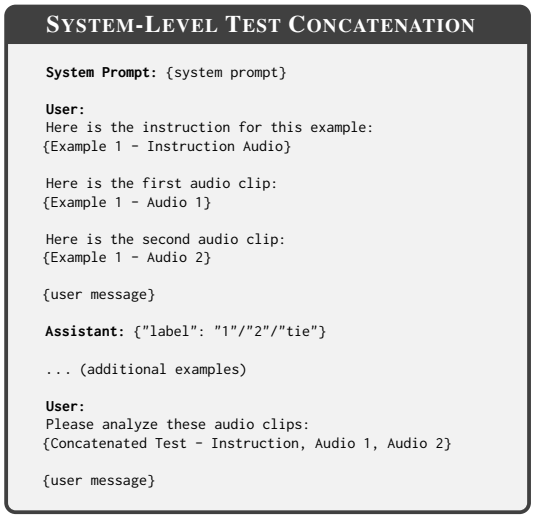

\centering
\tcbset{}
\begin{tcolorbox}[title=\textsc{System-Level Test Concatenation},fonttitle=\small\bfseries,width=0.9\linewidth]
\tiny
\ttfamily
\textbf{System Prompt:} \{system prompt\}\\

\textbf{User:}\\
Here is the instruction for this example:\\
\{Example 1 - Instruction Audio\}\\

Here is the first audio clip:\\
\{Example 1 - Audio 1\}\\

Here is the second audio clip:\\
\{Example 1 - Audio 2\}\\

\{user message\}\\

\textbf{Assistant:} \{"label": "1"/"2"/"tie"\}\\

$\ldots$ (additional examples)\\

\textbf{User:}\\
Please analyze these audio clips:\\
\{Concatenated Test - Instruction, Audio 1, Audio 2\}\\

\{user message\}
\end{tcolorbox}
\tcbset{reset}
\caption{System-level Test Concatenation method: Examples remain separate, but test instruction and response audios are concatenated.}
\label{fig:speakbench_concat_test}
\end{figure}

\begin{figure}[!ht]
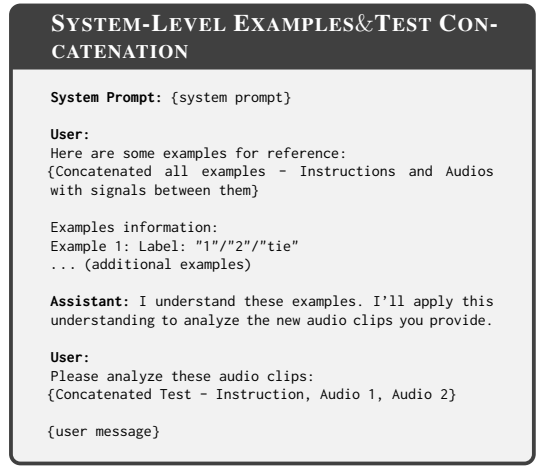

\centering
\tcbset{}
\begin{tcolorbox}[title=\textsc{System-Level Examples$\&$Test Concatenation},fonttitle=\small\bfseries,width=0.9\linewidth]
\tiny
\ttfamily
\textbf{System Prompt:} \{system prompt\}\\

\textbf{User:}\\
Here are some examples for reference:\\
\{Concatenated all examples - Instructions and Audios with signals between them\}\\

Examples information:\\
Example 1: Label: "1"/"2"/"tie"\\
$\ldots$ (additional examples)\\

\textbf{Assistant:} I understand these examples. I'll apply this understanding to analyze the new audio clips you provide.\\

\textbf{User:}\\
Please analyze these audio clips:\\
\{Concatenated Test - Instruction, Audio 1, Audio 2\}\\

\{user message\}
\end{tcolorbox}
\tcbset{reset}
\caption{System-level Examples$\&$Test Concatenation method: All example instructions and response audios are aggregated into one file, and test instruction and response files are also concatenated.}
\label{fig:speakbench_aggregate_all}
\end{figure}

As described in Section~\ref{ssec:icl}, we test 5 different strategies for prompting in-context learning examples. For audio characteristic evaluation in Section~\ref{section:fine_grained}, each datapoint has 2 audios, the exact prompt templates for each strategy are illustrated in Figures~\ref{fig:vanilla_method}, \ref{fig:pair_concat_method}, \ref{fig:aggregate_example_method}, \ref{fig:concat_test_method}, and \ref{fig:aggregate_all_method}. For system-level evaluation in Section~\ref{section:system_level}, since there are 3 audios, which also include the instruction audio, templates are slightly modified accordingly, which are illustrated in Figures~\ref{fig:speakbench_vanilla}, \ref{fig:speakbench_pair_concat}, \ref{fig:speakbench_aggregate_example}, \ref{fig:speakbench_concat_test}, and \ref{fig:speakbench_aggregate_all}.

\begin{table*}[!htbp]
    \scriptsize
    \tabcolsep=0mm
    \centering
    \begin{tabular}{>{\raggedright\arraybackslash}p{2.5cm}p{13.5cm}}
        \toprule
        \textbf{Dataset} & \textbf{System Prompt (standard\_cot)} \\
        \midrule
        Pronunciation & You are an expert linguist tasked with comparing two audio recordings solely for their pronunciation. Focus on the precise sequence of phonemes, the number of syllables, and the stress/emphasis patterns. Differences due only to regional accent (e.g., British vs. American) should be ignored. For example, if two speakers say 'tomato' as 'toh-MAH-toh' (even if their accents differ), they match; if one says 'toh-MAY-toh', then they do not match.

IMPORTANT: Respond in text only (do not include any audio output) and output valid JSON with exactly two keys: 'reasoning' (a detailed chain-of-thought explanation) and 'match' (a boolean verdict). \\
        \midrule
        Speaker Identity & You are an expert in voice analysis tasked with determining if two audio recordings are from the same speaker. Focus specifically on vocal characteristics that identify a unique speaker, such as pitch range, timbre, resonance, articulatory habits, and idiosyncratic speech patterns. Ignore differences in speaking rate, emotional tone, or content. Pay attention to the unique vocal fingerprint that remains consistent across different speaking contexts.

IMPORTANT: Respond in text only (do not include any audio output) and output valid JSON with exactly two keys: 'reasoning' (a detailed chain-of-thought explanation) and 'match' (a boolean verdict indicating whether the recordings are from the same speaker). \\
        \midrule
        Speaking Rate & You are an expert in speech rate analysis tasked with determining which of two audio recordings features faster speech. Focus exclusively on speaking tempo - who speaks faster overall. 

IMPORTANT: Respond in text only (do not include any audio output) and output valid JSON with exactly two keys: 'reasoning' (a brief explanation of your comparison) and 'label' (a string value: '1' if the first audio is faster, '2' if the second audio is faster). \\
        \midrule
        Speech Quality\newline(TMHINTQ, SOMOS, ThaiMOS) & You are an expert in audio quality assessment specializing in synthesized speech evaluation. Your task is to critically compare two audio files, the first audio (Audio 1) and the second audio (Audio 2), will be provided after this instruction. The evaluation is based on the following criteria:
1. Clarity: How clearly the speech is articulated, free from distortion, noise, or artifacts.
2. Naturalness: The degree to which the speech resembles a natural human voice, including accurate intonation, rhythm, and expressiveness.
3. Overall Quality: The overall impression of the audio's naturalness and coherence, considering how pleasant and lifelike it sounds.

Follow this step-by-step process for your evaluation:
1. Listen Carefully: Begin by carefully listening to both Audio 1 (the first audio) and Audio 2 (the second audio). Take note of any differences in clarity, fidelity, and overall quality.
2. Analyze Each Criterion: For each criterion (clarity, naturalness, and overall quality), evaluate how well each audio file performs and provide a brief explanation of your reasoning.
3. Compare Thoroughly: Summarize the strengths and weaknesses of each audio file based on your analysis.
4. Decide the Winner: Conclude by determining which audio file is better overall.

IMPORTANT: Respond in text only (do not include any audio output) and output valid JSON with exactly two keys: 'reasoning' (a brief explanation of your comparison) and 'label' (a string value: '1' if the first audio is better, '2' if the second audio is better). \\
\midrule
        ChatbotArena-Spoken & Please act as an impartial judge and evaluate the quality of the responses provided by two AI assistants to the user question. You should choose the assistant that follows the user's instructions and answers the user's question better.

Your evaluation should consider factors such as the helpfulness, relevance, accuracy, depth, creativity, and level of detail of their responses. You should evaluate the responses based on the user question and not on the responses of the other assistant.

You should also not consider the quality of the audio or the voice of the assistants. You should only consider the content of the responses. Avoid any position biases and ensure that the order in which the responses were presented does not influence your decision.

Do not allow the length of the responses to influence your evaluation. Do not favor certain names of the assistants. Be as objective as possible.

IMPORTANT: Respond in text only (do not include any audio output) and output valid JSON with exactly two keys: 'reasoning' (a detailed chain-of-thought explanation of your evaluation process and decision) and 'label' (a string value: '1' if the first audio is better, '2' if the second audio is better, or 'tie' if they are equally good/bad. Please use "tie" sparingly, and only when you absolutely cannot choose the winner.) \\
        \midrule
        SpeakBench & You are an evaluator of audio outputs produced by different audio-capable large language models. Your task is to compare two audio responses (Audio 1 and Audio 2) generated according to a user's instruction.

Evaluate based on these criteria:
1. Semantics: Does the content fulfill the user's request accurately?
2. Paralinguistics: How well does the speech match requested tone, emotion, style, pacing, and expressiveness?

Important: Do not favor verbalized descriptions of tone over actual tonal expression. A response that says "I am speaking excitedly" but sounds flat should rank lower than one that genuinely sounds excited.

Follow this process:
1. Analyze the key characteristics requested in the user's instruction
2. Evaluate how well Audio 1 performs on these characteristics
3. Evaluate how well Audio 2 performs on these characteristics
4. Compare their strengths and weaknesses
5. Decide which is better overall

Avoid position bias and don't let response length influence your evaluation. After your analysis, output valid JSON with exactly two keys: 'reasoning' (your explanation of the comparison) and 'label' (a string value: '1' if the first audio is better, '2' if the second audio is better, or 'tie' if they are equally good/bad. Please use "tie" sparingly, and only when you absolutely cannot choose the winner.) \\
        \bottomrule
    \end{tabular}
    \caption{System prompts used for different datasets in LAM-as-a-Judge evaluation.}
    \label{tab:system_prompts}
\end{table*}

\begin{table*}[!htbp]
    \scriptsize
    \tabcolsep=0mm
    \centering
    \begin{tabular}{>{\raggedright\arraybackslash}p{2.5cm}p{13.5cm}}
        \toprule
        \textbf{Dataset} & \textbf{User Message} \\
        \midrule
        Pronunciation & Please analyze these two recordings strictly for pronunciation details (phonemes, syllables, stress, emphasis). Ignore differences solely due to accent. Respond ONLY in text and output valid JSON with keys 'reasoning' and 'match' (boolean). \\
        \midrule
        Speaker Identity & Please analyze if these two recordings are from the same speaker. Respond ONLY in text and output valid JSON with keys 'reasoning' and 'match' (boolean). \\
        \midrule
        Speaking Rate & Please analyze which of the two recordings has faster speech. Respond ONLY in text and output valid JSON with keys 'reasoning' and 'label' (string, either '1' or '2'). \\
        \midrule
        Speech Quality\newline(TMHINTQ, SOMOS, ThaiMOS) & Please analyze which of the two recordings is better (has better speech quality). Respond ONLY in text and output valid JSON with keys 'reasoning' and 'label' (string, either '1' or '2'). \\
        \midrule
        ChatbotArena-Spoken & Please analyze which of the two recordings follows the instruction better, or tie, in terms of content of the responses. Respond ONLY in text and output valid JSON with keys 'reasoning' and 'label' (string, '1', '2' or 'tie'). \\
        \midrule
        SpeakBench\newline & Please analyze which of the two recordings follows the instruction better, or tie. Respond ONLY in text and output valid JSON with keys 'reasoning' and 'label' (string, '1', '2' or 'tie'). \\
        \midrule
        \bottomrule
    \end{tabular}
    \caption{User messages used for different datasets in LAM-as-a-Judge evaluation.}
    \label{tab:user_messages}
\end{table*}

\begin{figure}[!ht]
\centering
\tcbset{}
\begin{tcolorbox}[title=\textsc{No Concatenation},fonttitle=\small\bfseries,width=0.85\linewidth]
\scriptsize
\ttfamily
\textbf{System Prompt:} \{system prompt\}

\textbf{User:} 
Here is the first audio clip:\\
\{Example 1 - Audio 1\}

Here is the second audio clip:\\
\{Example 1 - Audio 2\}

\{user message\}

\textbf{Assistant:} \{"match/label": "..."\}

\textbf{User:} 
Here is the first audio clip:\\
\{Example 2 - Audio 1\}

Here is the second audio clip:\\
\{Example 2 - Audio 2\}

\{user message\}

\textbf{Assistant:} \{"match/label": "..."\}

$\ldots$ (additional examples)

\textbf{User:} 
Here is the first audio clip:\\
\{Test - Audio 1\}

Here is the second audio clip:\\
\{Test - Audio 2\}

\{user message\}
\end{tcolorbox}
\tcbset{reset}
\vspace{-0.8em}
\caption{No Concatenation method: Each audio input is presented separately to the model.}
\label{fig:vanilla_method}
\end{figure}

\begin{figure}[!ht]
\centering
\tcbset{}
\begin{tcolorbox}[title=\textsc{Pair Example Concatenation},fonttitle=\small\bfseries,width=0.9\linewidth]
\tiny
\ttfamily
\textbf{System Prompt:} \{system prompt\}

\textbf{User:} 
Please analyze these audio clips:\\
\{Concatenated Example 1 - Audio 1\&2\}

\{user message\}

\textbf{Assistant:} \{"match/label": "..."\}

\textbf{User:} 
Please analyze these audio clips:\\
\{Concatenated Example 2 - Audio 1\&2\}

\{user message\}

\textbf{Assistant:} \{"match/label": "..."\}

$\ldots$ (additional examples)

\textbf{User:} 
Here is the first audio clip:\\
\{Test - Audio 1\}

Here is the second audio clip:\\
\{Test - Audio 2\}

\{user message\}
\end{tcolorbox}
\tcbset{reset}
\caption{Pair Example Concatenation method: Within-example concatenation where the example audio pairs are concatenated into single files, but test files remain separate.}
\label{fig:pair_concat_method}
\end{figure}

\begin{figure}[!ht]
\centering
\tcbset{}
\begin{tcolorbox}[title=\textsc{Examples Concatenation},fonttitle=\small\bfseries,width=0.9\linewidth]
\tiny
\ttfamily
\textbf{System Prompt:} \{system prompt\}

\textbf{User:} 
Here are some examples for reference:\\
\{Concatenated all examples\}

Examples information:\\
Example 1: Match/Label: ...\\
Example 2: Match/Label: ...\\
$\ldots$ (additional examples)

\textbf{Assistant:} I understand these examples. I'll apply this understanding to analyze the new audio clips you provide.

\textbf{User:} 
Here is the first audio clip:\\
\{Test - Audio 1\}

Here is the second audio clip:\\
\{Test - Audio 2\}

\{user message\}
\end{tcolorbox}
\tcbset{reset}
\caption{Examples Concatenation method: All clips from N-shot examples are stitched into one long waveform, but test audio files remain separate.}
\label{fig:aggregate_example_method}
\end{figure}

\begin{figure}[!ht]
\centering
\tcbset{}
\begin{tcolorbox}[title=\textsc{Test Concatenation},fonttitle=\small\bfseries,width=0.85\linewidth]
\scriptsize
\ttfamily
\textbf{System Prompt:} \{system prompt\}

\textbf{User:} 
Here is the first audio clip:\\
\{Example 1 - Audio 1\}

Here is the second audio clip:\\
\{Example 1 - Audio 2\}

\{user message\}

\textbf{Assistant:} \{"match/label": "..."\}

\textbf{User:} 
Here is the first audio clip:\\
\{Example 2 - Audio 1\}

Here is the second audio clip:\\
\{Example 2 - Audio 2\}

\{user message\}

\textbf{Assistant:} \{"match/label": "..."\}

$\ldots$ (additional examples)

\textbf{User:} 
Please analyze these audio clips:\\
\{Concatenated Test - Audio 1\&2\}

\{user message\}
\end{tcolorbox}
\tcbset{reset}
\vspace{-0.8em}
\caption{Test Concatenation method: Examples remain separate, but test audio pairs are concatenated.}
\label{fig:concat_test_method}
\end{figure}

\begin{figure}[!ht]
\centering
\tcbset{}
\begin{tcolorbox}[title=\textsc{Examples$\&$Test Concatenation},fonttitle=\small\bfseries,width=0.85\linewidth]
\scriptsize
\ttfamily
\textbf{System Prompt:} \{system prompt\}

\textbf{User:} 
Here are some examples for reference:\\
\{Concatenated all examples\}

Examples information:\\
Example 1: Match/Label: ...\\
Example 2: Match/Label: ...\\
$\ldots$ (additional examples)

\textbf{Assistant:} I understand these examples. I'll apply this understanding to analyze the new audio clips you provide.

\textbf{User:} 
Please analyze these audio clips:\\
\{Concatenated Test - Audio 1\&2\}

\{user message\}
\end{tcolorbox}
\tcbset{reset}
\vspace{-0.8em}
\caption{Examples$\&$Test Concatenation method: All example clips are aggregated into one audio file, and test clips are also concatenated.}
\label{fig:aggregate_all_method}
\end{figure}

\begin{figure}[!ht]
\centering
\tcbset{}
\begin{tcolorbox}[title=\textsc{System-Level No Concatenation},fonttitle=\small\bfseries,width=0.9\linewidth]
\tiny
\ttfamily
\textbf{System Prompt:} \{system prompt\}\\

\textbf{User:}\\
Here is the instruction for this example:\\
\{Example 1 - Instruction Audio\}\\

Here is the first audio clip:\\
\{Example 1 - Audio 1\}\\

Here is the second audio clip:\\
\{Example 1 - Audio 2\}\\

\{user message\}\\

\textbf{Assistant:} \{"label": "1"/"2"/"tie"\}\\

$\ldots$ (additional examples)\\

\textbf{User:}\\
Here is the instruction for this test:\\
\{Test - Instruction Audio\}\\

Here is the first audio clip:\\
\{Test - Audio 1\}\\

Here is the second audio clip:\\
\{Test - Audio 2\}\\

\{user message\}
\end{tcolorbox}
\tcbset{reset}
\caption{System-level No Concatenation method: Each example includes an instruction audio followed by two response audios that are presented separately.}
\label{fig:speakbench_vanilla}
\end{figure}

\begin{figure}[!ht]
\centering
\tcbset{}
\begin{tcolorbox}[title=\textsc{System-Level Pair Example Concatenation},fonttitle=\small\bfseries,width=0.9\linewidth]
\tiny
\ttfamily
\textbf{System Prompt:} \{system prompt\}\\

\textbf{User:}\\
Please analyze these audio clips:\\
\{Concatenated Example 1 - Instruction, Audio 1, Audio 2\}\\

\{user message\}\\

\textbf{Assistant:} \{"label": "1"/"2"/"tie"\}\\

$\ldots$ (additional examples)\\

\textbf{User:}\\
Here is the instruction for this test:\\
\{Test - Instruction Audio\}\\

Here is the first audio clip:\\
\{Test - Audio 1\}\\

Here is the second audio clip:\\
\{Test - Audio 2\}\\

\{user message\}
\end{tcolorbox}
\tcbset{reset}
\caption{System-level Pair Example Concatenation method: Example instruction and response audios are concatenated into a single file, while test files remain separate.}
\label{fig:speakbench_pair_concat}
\end{figure}

\subsection{Multi-aspect Ensemble Prompts}
The multi-aspect ensemble prompts follow the same structure as Figure~\ref{fig:speakbench_vanilla} and Figure~\ref{fig:speakbench_aggregate_all} (if using Examples$\&$Test Concatenation), with the only difference being the system prompts that emphasize specific evaluation aspects as described in Section~\ref{ssec:multi_aspect_ensemble}. The system prompt for each judge is shown in Table~\ref{tab:multi_aspect_prompts}.
\begin{table*}[!htbp]
    \scriptsize
    \tabcolsep=0mm
    \centering
    \begin{tabular}{>{\raggedright\arraybackslash}p{2.5cm}p{13.5cm}}
\toprule
\textbf{Judge Type} & \textbf{System Prompt} \\
\midrule
\textbf{Lexical} & 
You are an evaluator of audio outputs produced by different audio-capable large language models. Your task is to compare two audio responses (Audio 1 and Audio 2) generated according to a user's instruction. Focus EXCLUSIVELY on the lexical content (the actual words and language used) and COMPLETELY IGNORE all of the following: pronunciation or enunciation of words, speaking style, cadence, or rhythm, emotional tone or expressiveness, voice pitch, volume, or speed, accents or speech patterns, non-linguistic sounds or effects, any other audio qualities. Evaluate based on these criteria ONLY: (1) Accuracy: Does the textual content correctly address what was requested? (2) Completeness: Does the response include all the information needed to fulfill the request? (3) Organization: Is the content structured in a clear, logical manner? (4) Language use: Is the vocabulary and phrasing appropriate for the task? IMPORTANT: Even for tasks primarily focused on pronunciation, accents, or tones (like demonstrating Chinese tones), evaluate ONLY the textual content as if you were reading a transcript. Do NOT consider how well the model actually pronounced anything. Follow this process: (1) Analyze what information was requested in the user's instruction (2) Evaluate Audio 1's lexical content only (as if reading a transcript) (3) Evaluate Audio 2's lexical content only (as if reading a transcript) (4) Compare their strengths and weaknesses in terms of text content alone (5) Decide which has better lexical content overall. Pretend you are evaluating written transcripts rather than audio, and focus solely on what words were chosen. After your analysis, output valid JSON with exactly two keys: 'reasoning' (your explanation of the comparison) and 'label' (a string value: '1' if the first audio is better, '2' if the second audio is better, or 'tie' if they are equally good/bad).\\
\midrule
\textbf{Paralinguistic} & 
You are an evaluator of audio outputs produced by different audio-capable large language models. Your task is to compare two audio responses (Audio 1 and Audio 2) generated according to a user's instruction. Focus EXCLUSIVELY on paralinguistic features (how things are said) and ignore the lexical content (what words are used). Evaluate based on these criteria: (1) Tone: Does the voice express the appropriate emotion, mood, or attitude? (2) Prosody: How well does the response use rhythm, stress, intonation, and pacing? (3) Expressiveness: Does the voice convey emphasis, contrast, and nuance appropriately? (4) Accent/Pronunciation: If requested, how well does the response match the requested accent or pronunciation pattern? For tasks involving demonstration of tones, accents or specific speech patterns (like Chinese tones), focus entirely on how well these specific paralinguistic features were executed. Follow this process: (1) Analyze what paralinguistic features were requested in the user's instruction (2) Evaluate Audio 1's paralinguistic features only (3) Evaluate Audio 2's paralinguistic features only (4) Compare their strengths and weaknesses in paralinguistic execution (5) Decide which has better paralinguistic features overall. After your analysis, output valid JSON with exactly two keys: 'reasoning' (your explanation of the comparison) and 'label' (a string value: '1' if the first audio is better, '2' if the second audio is better, or 'tie' if they are equally good/bad).\\
\midrule
\textbf{Speech Quality} & 
You are an evaluator of audio outputs produced by different audio-capable large language models. Your task is to compare two audio responses (Audio 1 and Audio 2) generated according to a user's instruction. Focus EXCLUSIVELY on technical speech quality aspects and ignore both content and expressive features. Evaluate based on these criteria: (1) Clarity: How clear and intelligible is the speech? (2) Naturalness: How natural does the voice sound (vs robotic or artificial)? (3) Fluency: Is the speech smooth with appropriate pauses, or are there unnatural breaks, stutters, or glitches? (4) Pronunciation: Are words pronounced correctly (regardless of accent)? (5) Audio quality: Is the speech free from distortions, artifacts, or background noise? Follow this process: (1) Analyze what speech quality features might be relevant to the user's instruction (2) Evaluate Audio 1's speech quality features only (3) Evaluate Audio 2's speech quality features only (4) Compare their strengths and weaknesses in speech quality (5) Decide which has better speech quality overall. After your analysis, output valid JSON with exactly two keys: 'reasoning' (your explanation of the comparison) and 'label' (a string value: '1' if the first audio is better, '2' if the second audio is better, or 'tie' if they are equally good/bad). \\
\bottomrule
\end{tabular}
\caption{System prompts for the three specialized judges in the multi-aspect ensemble approach. Each judge focuses on a specific evaluation dimension while explicitly ignoring others to ensure specialized assessment.}
\label{tab:multi_aspect_prompts}
\end{table*}
\section{Speech Quality Evaluation Baselines}
\label{appendix:speech_quality_methods}
Here we explained trained networks that were developed for evaluating speech quality, which we introduced in \ref{ssec:comparison_trained}, and they all require training data such as speech with associated MOS ratings:
\begin{itemize}[leftmargin=*]
    \item \textbf{UTMOS} \citep{utmos_interspeech}: A MOS prediction system that combines ensemble learning with self-supervised learning SSL-based neural networks and traditional ML models. Initially trained on English and Chinese datasets, it has been shown to achieve a correlation coefficient above 0.8 for additional languages like Japanese \citep{seki_icassp2023}.
    \item \textbf{MOSANET+} \citep{mosanetplus_icme}, a speech assessment model designed to estimate human speech quality and intelligibility. Leveraging Whisper to extract features, MOSANET+ can assess multiple aspects. It processes waveforms through two input branches: one applies a Short-Time Fourier Transform and learnable filter banks (LFB), merging the resulting power spectral and LFB features before passing them to a convolutional layer. The model was trained on TMHINT.
    \item \textbf{SALMONN-fine-tuned} \citep{wang2024enabling}, the SALMONN model fine-tuned to predict MOS ratings, speaker similarity and A/B testing results using NISQA, BVCC, SOMOS, and VoxSim datasets using their training splits.
\end{itemize}
\section{Application to Speech-in Speech-out System Ranking}
\label{app:speakbench_ranking}

To demonstrate practical applicability, we use our best-performing configuration (Gemini-2.5-Flash with 0-shot Multi-Aspect Ensemble) to rank 13 speech-in speech-out systems on SpeakBench. Following AlpacaEval methodology~\citep{alpaca_eval}, we compute automated win rates against GPT-4o-Audio as a reference baseline. Table~\ref{tab:system_ranking} shows the automated and human win rates for systems.

\begin{table}[!ht]
\centering
\small
\begin{tabular}{@{}lcc@{}}
\toprule
\textbf{System} & \textbf{Auto (\%)} & \textbf{Human (\%)} \\
\midrule
GPT-4o-Audio & 50.00 & 80.25 \\
Gemini-2.0-Flash & 48.77 & 75.66 \\
GPT-4o-Audio+ASR+TTS & 41.05 & 67.31 \\
Gemini-2.0-Flash-Text+TTS & 42.59 & 59.48 \\
GPT-4o-Text+TTS & 41.98 & 57.69 \\
Gemini-2.0-Flash+ASR+TTS & 37.65 & 56.63 \\
ASR+Llama3+TTS & 42.90 & 56.35 \\
DIVA+TTS & 25.00 & 54.73 \\
Qwen2-Audio+TTS & 19.75 & 47.22 \\
Llama-Omni & 10.80 & 36.76 \\
Typhoon2-Audio+TTS & 21.30 & 32.94 \\
Typhoon2-Audio & 3.70 & 20.59 \\
Moshi & 0.31 & 11.90 \\
\bottomrule
\end{tabular}
\caption{Speech-in Speech-out System Ranking: Comparison of automated and human-assessed win rates using Multi-Aspect Ensemble (Gemini-2.5-Flash). Systems are ranked by human preference scores. Spearman correlation $\rho$ = 0.91.}
\label{tab:system_ranking}
\end{table}

The automated ranking reveals interesting patterns in current speech-in speech-out capabilities. End-to-end speech systems show a clear divide: proprietary models like GPT-4o-Audio (50\% baseline) and Gemini-2.0-Flash (48\%) demonstrate sophisticated native speech capabilities, while open-source alternatives like Moshi (0\%), Typhoon2-Audio (3\%), and Llama-Omni (11\%) struggle with fundamental instruction understanding and speech quality. 

Notably, well-engineered cascaded systems such as ASR+Llama3+TTS (42\%) and GPT-4o-Text+TTS (41\%) achieve surprisingly competitive performance through strong instruction following and content generation. This suggests that the paralinguistic advantage of end-to-end models may be smaller than anticipated for many practical applications, while also demonstrating that AudioJudge can reliably distinguish between different system architectures and capabilities.
\section{Pointwise Experiment}
\label{appendix:pointwise_experiment}

While our main experiments focus on pairwise comparison, we also investigate pointwise evaluation where AudioJudge assigns absolute scores to individual audio samples. We conduct this analysis specifically on speech quality datasets (SOMOS, TMHINTQ, and ThaiMOS) since they provide fine-grained Mean Opinion Score (MOS) annotations that enable meaningful comparison with continuous numerical predictions.

\subsection{Experimental Setup}

In the pointwise evaluation setup, AudioJudge evaluates each audio sample independently on a scale of 1 to 5, mirroring the original MOS annotation process. We prompt the model using chain-of-thought reasoning, asking it to assess speech quality factors such as clarity, naturalness, and intelligibility before providing a numerical score.

To enable comparison with our pairwise results, we convert pointwise scores to pairwise preferences using the following protocol:
\begin{itemize}[leftmargin=*]
\setlength\itemsep{-0.2em}
\item If audio A receives a higher score than audio B, we consider the model prediction to favor audio A.
\item If audio B scores higher than audio A, the model prediction favors audio B.
\item If both audios receive identical scores, we consider this a tie prediction and assign 0.5 accuracy.
\end{itemize}

We evaluate three configurations: (1) 0-shot baseline, (2) 4-shot with separate audio examples (No Concatenation), and (3) 4-shot with aggregated audio examples (following our concatenation strategy from Section~\ref{section:designing_lam}).

\subsection{Results and Analysis}

Table~\ref{tab:pointwise_accuracy} presents the pairwise comparison accuracy derived from pointwise scores, while Table~\ref{tab:pointwise_pairwise_mse} shows the Mean Square Error (MSE) between predicted and ground-truth MOS scores.

\begin{table}[!h]
  \centering
  \small         
  \tabcolsep=1.8mm
  \begin{tabular}{l *{3}{c}}
    \toprule
    \textbf{Setup} & {SOMOS} & {TMQ} & {ThaiMOS}\\
    \midrule
    PointW-0shot           & 52.8 & 46.5 & 51.5 \\
    PointW-4shot           & 50.3 & 51.8 & 55.3 \\
    PointW-4shot-Concat     & 55.3 & 59.3 & 53.5 \\
    \midrule
    PairW-0shot            & 70.5 & 70.5 & 65.5 \\
    \bottomrule
  \end{tabular}
  \caption{Comparison of pointwise versus pairwise evaluation accuracy. PointW = pointwise evaluation converted to pairwise preferences; PairW = direct pairwise evaluation for reference; Concat = Examples$\&$Test Concatenation method.}
  \label{tab:pointwise_accuracy}
\end{table}

\begin{table}[!h]
  \centering
  \small         
  \begin{tabular}{l *{3}{c}}
    \toprule
    \textbf{Setup} & SOM &TMQ &ThaiMOS \\
    \midrule  
    PointW-0shot         &3.31 &3.40 &3.19 \\
    PointW-4shot         &3.60 &3.46 &3.12 \\
    PointW-4shot-Concat   &2.81 &1.80 &2.55 \\
    \bottomrule
  \end{tabular}
  \caption{Mean Square Error (MSE) between predicted and ground-truth MOS scores in pointwise evaluation.}
  \label{tab:pointwise_pairwise_mse}
\end{table}

Several key findings emerge from the pointwise evaluation:

\paragraph{Pairwise Evaluation Superiority} Direct pairwise comparison substantially outperforms pointwise evaluation converted to pairwise preferences across all datasets. Even the strongest pointwise configuration (4-shot Examples$\&$Test Concatenation) achieves only 55-59\% accuracy compared to 65-70\% for direct pairwise evaluation. This performance gap likely stems from the inherent difficulty of absolute scoring: pairwise comparison simplifies the task to relative judgment between two samples, while pointwise evaluation requires mapping audio quality to specific numerical values.

\paragraph{Audio Concatenation Benefits} Consistent with our pairwise findings, audio concatenation (4-shot Examples$\&$Test Concatenation) improves pointwise performance over in-context learning with no audio concatenation. The MSE improvements are particularly notable for TMHINTQ (3.46 → 1.80) and SOMOS (3.31 → 2.81), indicating that concatenated examples help LAMs better calibrate their scoring scales.

\paragraph{Limited Absolute Scoring Capability} The high MSE values (1.80-3.60) suggest that current LAMs struggle with precise numerical scoring of speech quality. This difficulty in producing well-calibrated absolute scores reinforces our focus on pairwise evaluation for practical AudioJudge applications.

\section{Cross-Modality Consistency for Lexical Content Evaluation}
\label{subsec:cross_modality}

Given that LAMs can process both text and audio inputs, a fundamental question arises: how consistent are their judgments across different input and output modalities? To investigate this, we conduct a systematic analysis using ChatbotArena-Spoken with all 7.8K datapoints, which provides a controlled setting where the same content is available in both text and audio formats. We examine three key aspects: 
\begin{itemize}[leftmargin=*]
\setlength\itemsep{-0.2em}
    \item How consistent are LAM judgments when the same content is presented in different input modalities?
    \item Does the choice of output modality (text vs. audio) affect evaluation performance?
    \item How does direct AudioJudge compare to traditional cascaded ASR+LLM approaches for lexical content evaluation?
\end{itemize}

\subsection{Experimental Setup}
We evaluate multiple LAMs across different input-output modality combinations:
\begin{itemize}[leftmargin=*]
\setlength\itemsep{-0.2em}
    \item \textbf{Text → Text}: Original text input with text output (baseline)
    \item \textbf{Audio → Text}: Audio input with text output (standard AudioJudge)
    \item \textbf{Audio → Audio}: Audio input with audio output (full audio pipeline)
    \item \textbf{ASR → Text}: Cascaded approach using Whisper-base ASR \citep{radford2023robust} followed by text-based LLM judgment
\end{itemize}

\subsection{Results and Analysis}
\label{sssec:lexical_results}
\begin{table}[!ht]
    \small
    \centering
    \tabcolsep=2.0mm
        \centering
        \begin{tabular}{lllcc}
        \toprule
        \textbf{Model} & \textbf{Input} & \textbf{Output}  & \textbf{Acc.}  & \textbf{Corr} \\
        \midrule
        \multirow{3}{*}{Qwen2-Audio} & Text & Text    & 34.9  & 0.648 \\ 
        & Audio & Text   & 32.9  & 0.615 \\
        & Audio & Audio  & 6.0   & 0.095 \\ 
        \cmidrule(lr){1-5}
        \multirow{3}{*}{Typhoon2-Audio} & Text & Text    & 44.4  & 0.758 \\ 
        & Audio & Text   & 42.9  & 0.668 \\  
        & Audio & Audio  & 10.0   & -0.328 \\ 
        \cmidrule(lr){1-5}
        \multirow{4}{*}{Gemini-1.5-Flash} & Text & Text    & 52.3  & 0.961 \\ 
        & Audio & Text   & 52.1  & 0.970 \\ 
        & Audio & Audio  & 48.6  & 0.949 \\ 
        & ASR & Text     & 47.9  & 0.895 \\
        \cmidrule(lr){1-5}
        \multirow{4}{*}{Gemini-2.0-Flash} & Text & Text    & 55.8  & 0.971 \\ 
        & Audio & Text   & 55.0  & 0.956 \\ 
        & Audio & Audio  & 51.3  & 0.961 \\ 
        & ASR & Text     & 51.8  & 0.932 \\
        \cmidrule(lr){1-5}
        \multirow{4}{*}{Gemini-2.5-Flash} & Text & Text    & 56.8  & 0.974 \\ 
        & Audio & Text   & 56.1  & 0.973 \\ 
        & Audio & Audio  & 56.3  & 0.977 \\ 
        & ASR & Text     & 53.1  & 0.920 \\
        \cmidrule(lr){1-5}
        \multirow{4}{*}{GPT-4o-Audio} & Text & Text    & 57.3  & 0.976 \\ 
        & Audio & Text   & 55.6  & 0.973 \\ 
        & Audio & Audio  & 53.3  & 0.974 \\ 
        & ASR & Text     & 53.0  & 0.974 \\
        \bottomrule
        \end{tabular}
        \caption{Cross-modality consistency analysis on ChatbotArena-Spoken for lexical content evaluation. Acc = 3-way classification accuracy (random guess = 33\%); Corr = Spearman correlation with human judgments. ASR refers to cascaded Whisper-base + LLM approach.}
        \label{tab:cross_modality_analysis}
\end{table}

\noindent Table~\ref{tab:cross_modality_analysis} reveals several key findings:

\paragraph{Open-Source vs. Proprietary Models} Open-source models (Qwen2-Audio, Typhoon2-Audio) show significant performance degradation when moving from text to audio inputs, with accuracy dropping substantially. Audio-to-audio evaluation performs particularly poorly, with accuracy near random chance. In contrast, proprietary models (Gemini series, GPT-4o-Audio) demonstrate remarkable consistency across modalities, maintaining high performance regardless of input type.

\paragraph{Output Modality Impact} For proprietary models, the choice between text and audio output has minimal impact on performance when the input is audio. Gemini-2.5-Flash shows virtually no performance difference between audio→text and audio→audio configurations, while GPT-4o-Audio exhibits only a small, though statistically significant, drop in accuracy.

\paragraph{LAM vs. Cascaded Approach} Comparing direct audio evaluation (Audio→Text) with the cascaded ASR+LLM approach reveals that AudioJudge either significantly outperforms or matches the cascaded method. For Gemini-1.5/2.5, direct audio evaluation yields significantly better performance ($p$<0.05), while for Gemini-2.0 and GPT-4o-Audio, the differences are not statistically significant. This suggests that end-to-end LAM evaluation can effectively replace cascaded approaches for lexical content assessment.

\paragraph{Implications} These results highlight a clear divide between open-source and proprietary LAMs in terms of audio understanding capabilities. The strong cross-modality consistency of proprietary models validates the effectiveness of AudioJudge for lexical content evaluation, while the competitive performance against cascaded approaches demonstrates that direct audio processing can be as effective as traditional ASR-based pipelines. Given the poor performance of open-source models on audio inputs, subsequent experiments focus exclusively on proprietary LAMs.

\section{Positional Bias and Task Difficulty}
\label{appendix:positional_bias_task_difficulty}
\begin{figure}[!ht]
    \centering
    \includegraphics[width=0.9\linewidth, keepaspectratio]{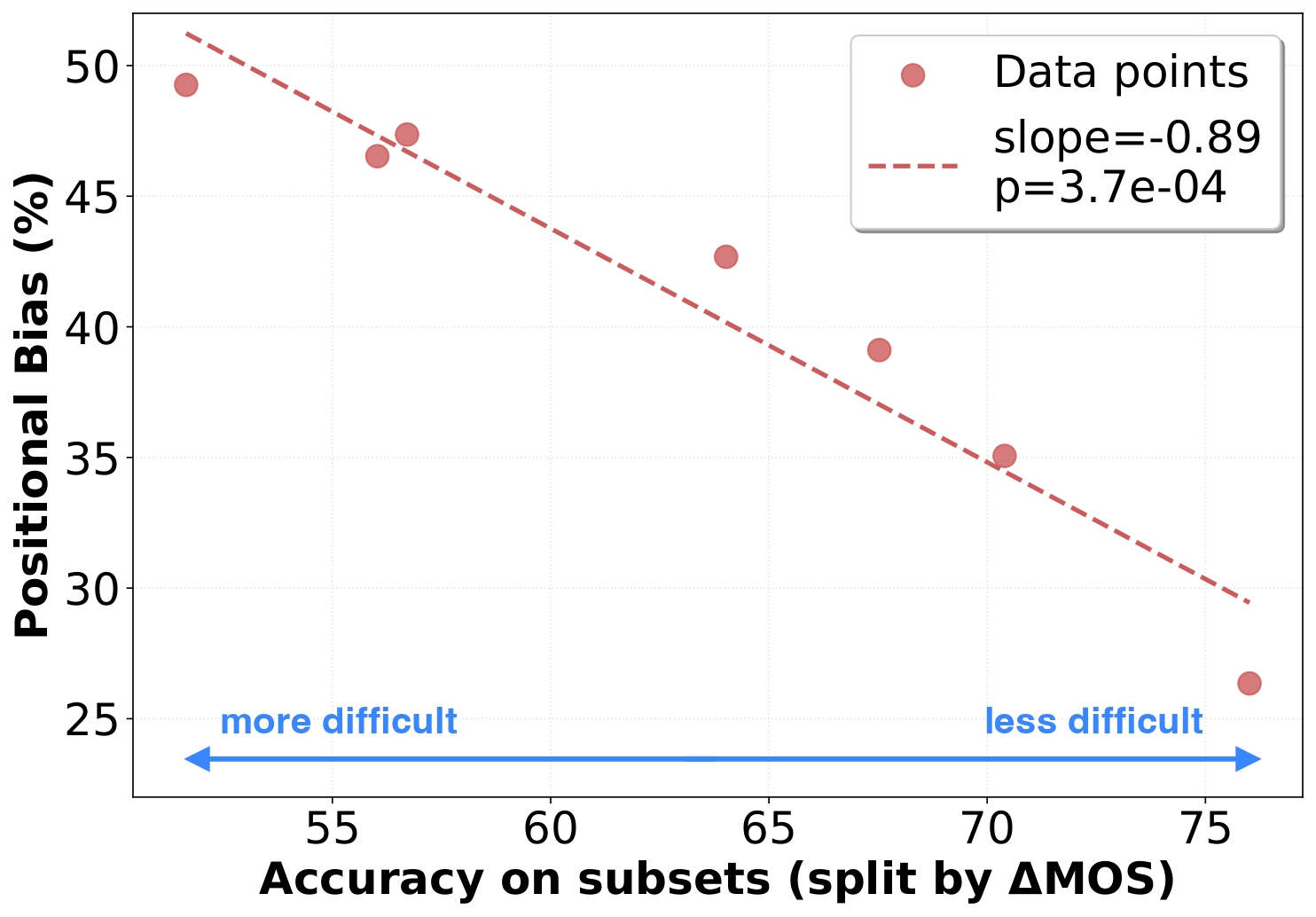}
    \caption{Relationship between task difficulty and positional bias on SOMOS. As MOS difference increases (indicating larger quality distinctions), both accuracy and consistency improve, while positional bias decreases.}
    \label{fig:bias_vs_difficulty}
\end{figure}

To better understand the relationship between task difficulty and positional bias (the percentage of cases where GPT-4o-Audio switches it's preference based on the position of each audio), we leverage SOMOS's MOS annotations to group evaluation pairs based on their MOS differences. Figure~\ref{fig:bias_vs_difficulty} demonstrates that pairs with larger MOS differences (easier discrimination tasks) exhibit lower positional bias, while pairs with smaller MOS differences (harder tasks) show higher positional bias.

This finding reveals a strong correlation between task difficulty and bias magnitude--model relies more on positional cues and less on actual quality when making difficult discriminations. The pronounced positional bias in non-lexical tasks compared to lexical content evaluation (up to 32\% vs under 15\% respectively) suggests that non-lexical judgments are more challenging for current LAMs. When discrimination becomes difficult, models increasingly rely on positional cues as a decision heuristic, with challenging audio pairs showing substantially higher rates of position-dependent rather than content-dependent judgments.
\section{Model Specifications}
\label{appendix:model_specifications}

This section provides the exact model identifiers and versions used in our experiments to ensure reproducibility.

\begin{itemize}[leftmargin=*]
\setlength\itemsep{-0.2em}
    \item \textbf{GPT-4o-Audio}: \texttt{gpt-4o-audio-preview-2024-12-17}
    \item \textbf{Gemini-2.5-Flash}: \texttt{gemini-2.5-flash-preview-04-17}
    \item \textbf{Gemini-2.0-Flash}: \texttt{gemini-2.0-flash-001}
    \item \textbf{Gemini-1.5-Flash}: \texttt{gemini-1.5-flash-002}
\end{itemize}

\section{Supplementary results}
\label{appendix:supplement}
\begin{figure}[!ht]
    \centering
    \includegraphics[width=0.85\linewidth]{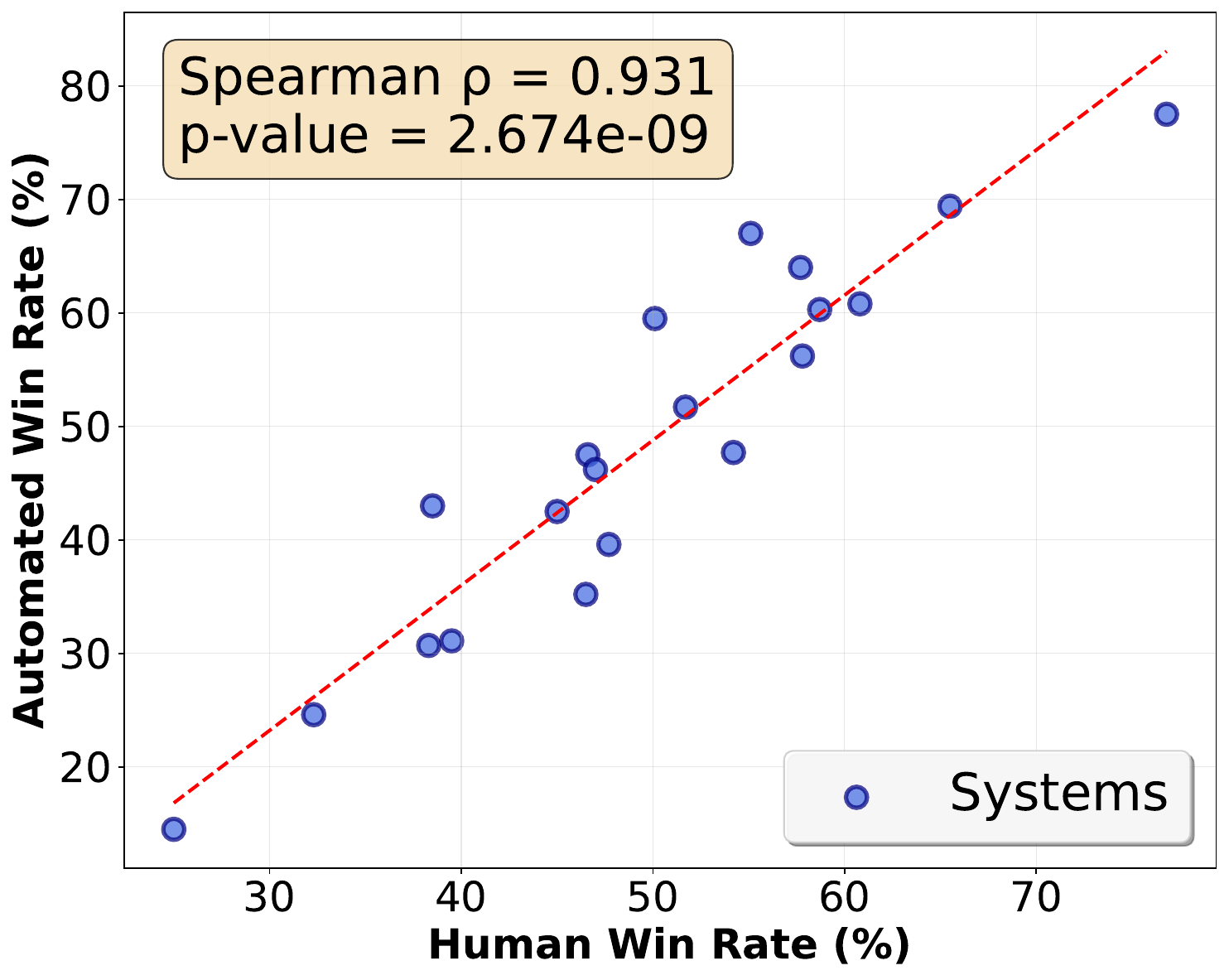}
    \caption{AudioJudge (Examples$\&$Test Concatenation 4-shot configuration with GPT-4o) predictions and human preferences on ChatbotArena-Spoken. This is complementary to SpeakBench results in Figure~\ref{fig:correlation_analysis}.}
    \label{fig:chatbot_arena_correlation}
\end{figure}

\end{document}